\documentclass[final]{article}
\usepackage[utf8]{inputenc}
\PassOptionsToPackage{numbers}{natbib}
\usepackage{neurips_2021}
\usepackage{amsmath}
\usepackage{indentfirst}
\usepackage{subfig}
\usepackage{courier}
\usepackage{graphicx}
\usepackage{algorithm,algorithmic}
\usepackage{dsfont}
\usepackage{color}
\usepackage{multirow}
\usepackage{verbatim}
\usepackage{siunitx}
\usepackage{longtable,tabularx}
\usepackage{threeparttable}
\usepackage{makecell}
\usepackage{epstopdf}

\usepackage{multirow}
\usepackage{hhline}
\usepackage{textcomp}
\usepackage{gensymb}
\usepackage{url}
\usepackage[normalem]{ulem}
\usepackage{xcolor}         

\newcommand{\beq}{\begin{eqnarray}}
\newcommand{\eeq}{\end{eqnarray}}
\newcommand{\bal}{\begin{eqnarray}\begin{aligned}}
\newcommand{\eal}{\end{aligned}\end{eqnarray}}

\newcommand{\re}{\mathds{R}}

\newcommand{\RNum}[1]{\uppercase\expandafter{\romannumeral #1\relax}}
\newcommand{\eqr}[1]{Eq.~\eqref{#1}}

\newcommand{\inr}[1]{\in\mathds{R}^{{#1}}}
\newcommand{\mx}{\mathbf{x}}
\newcommand{\mb}{\mathbf{b}}

\newcommand{\mh}{\mathbf{h}}
\newcommand{\me}{\mathbf{e}}

\newcommand{\bu}{\mathbf{u}}
\newcommand{\mq}{\mathbf{q}}

\newcommand{\mB}{\mathbf{B}}
\newcommand{\mW}{\mathbf{W}}

\makeatletter
\newcommand\notsotiny{\@setfontsize\notsotiny{5.7}{6.7}}
\makeatother

\title{Conditionally-Parameterized, Discretization-Aware Neural Networks for Mesh-Based Modeling of Physical Systems}

\author{%
\vspace{-0.1cm}
Jiayang Xu \\
\texttt{davidxu@umich.edu} 
\And
Aniruddhe Pradhan\\
\texttt{anipra@umich.edu} 
\And
Karthik Duraisamy\\
\texttt{kdur@umich.edu} 
\And
\vspace{-0.1cm}
Department of Aerospace Engineering,
University of Michigan,
Ann Arbor, MI 48109.
}

\begin{document}
\maketitle
\vspace{-0.3cm}
\begin{abstract} 
Simulations of complex physical systems are typically realized by discretizing partial differential equations (PDEs) on unstructured meshes. While neural networks have recently been explored for surrogate and reduced order modeling of PDE solutions, they often ignore interactions or hierarchical relations between input features, and process them as concatenated mixtures. We generalize the idea of conditional parameterization -- using trainable functions of input parameters to generate the weights of a neural network, and extend them in a flexible way to encode critical information. Inspired by discretized numerical methods, choices of the parameters include physical quantities and mesh topology features. The functional relation between the modeled features and the parameters is built into the network architecture. The method is implemented on different networks and applied to frontier scientific machine learning tasks including the discovery of unmodeled physics, super-resolution of coarse fields, and the simulation of unsteady flows with chemical reactions. The results show that the conditionally-parameterized networks provide superior performance compared to their traditional counterparts. The CP-GNet - an architecture that can be trained on very few data snapshots - is proposed as the first deep learning model capable of standalone prediction of reacting flows on irregular meshes.
\end{abstract}
\vspace{-0.3cm}

\section{Introduction}

Numerical simulations of partial differential equations (PDEs) have become an indispensable tool in the study of complex physical systems. High-resolution simulations are, however, prohibitively expensive or intractable in many practical problems. Machine learning techniques have recently been explored to improve the efficiency and accuracy of traditional numerical methods. Successful applications include nonlinear model order reduction~\citep{guo2016convolutional,lee2018model, murata2020nonlinear}, model augmentation~\citep{singh2017machine,  wu2018physics, karthikPRF2021}, and super-resolution~\citep{fukami2021machine, pradhan2021variational, guo2020ssr}. Neural networks have also been used to replace traditional PDE-based solvers, and serve as a standalone prediction tool. Popular approaches include auto-regressive time-series predictions~\citep{xu2020multi, gonzalez2018deep, maulik2020reduced, mohan2019compressed, maulik2020time}, Physics-Informed Neural Networks (PINNs)~\citep{lagaris1998artificial, raissi2019physics, sun2020surrogate}. 

Despite promising results on canonical problems,  commonly used network architectures such as autoencoders and CNNs have inherent limitations. 
An autoencoder generates a fixed mapping between the  geometric coordinates and the encoded digits. This limits their portability for new geometries and dynamic patterns. A CNN requires an interpolation of existing data to a structured, Euclidean space, introducing additional cost and error. Irregular geometry boundaries require constructs such as elliptic coordinate transformation~\citep{gao2020phygeonet} and Signed Distance Function (SDF)~\citep{guo2016convolutional}. Moreover, models often ignore the hierarchical relations between heterogeneous features, and concatenate them into a single input vector, e.g. the common concatenation of the edge and node features in Graph Neural Networks (GNNs). The learning of high-order terms remains mostly unguided -- even simple quadratic terms are often fitted via a number of hidden units in a brute-force manner. 

With a focus on mesh-based modeling of physical systems, we use the idea of conditional parameterization (CP) to build the hierarchical relations between different physical quantities as well as numerical discretization information into the network architectures. The key contributions of our work are as follows~\footnote{The source code is released to facilitate future research at \url{https://github.com/davidxujiayang/cpnets}}:

1. We demonstrate that a drop-in CP modification can bring significant improvements for various existing models on several tasks essential to the modeling of physical systems.

2. We propose a conditionally parameterized graph neural network (CP-GNet), which effectively models complex physics such as chemical source terms, irregular mesh discretizations, and different types of boundary conditions.

3. We conduct extensive numerical tests and demonstrate state-of-the-art performances on problems of different complexities, ranging from the basic viscous Burgers equation to a complex reacting flow.



\section{Methodology}\label{sec overview}

{\bf Conditional Parametrization:}
The idea of conditional parametrization (CP) is to use trainable functions of input parameters to generate the weights of a neural network. To demonstrate this, we start from a standard dense (fully connected) layer:
\begin{equation}\label{eq standard}
\mathbf{h}(\mathbf{u}; \mathbf{W}, \mathbf{b}) = \sigma(\mathbf{Wu+b}),
\end{equation}
where $\bu\in\re^{n_x}$ is the input feature vector, $\mathbf{h}\in\re^{n_h}$ is the output hidden state vector, $\mathbf{W}\in\re^{n_h\times n_x}$ and $\mathbf{b}\in\re^{n_h}$ are the trainable weights and bias, and $\sigma$ is the activation function. It can be seen that in the evaluation stage, the values of $\mathbf{W}$ and $\mathbf{b}$ are fixed regardless of the inputs. Thus the performance of \eqr{eq standard} is largely limited by the interpolation range of training data.

By introducing a parameter vector $\mathbf{p}\in\re^{n_p}$ and a trainable function $f(\mathbf{p}):\re^{n_p}\rightarrow\re^{n_h\times n_x}$ that computes the weights $\mathbf{W}$ based on $\mathbf{p}$, the conditionally parameterized version of \eqr{eq standard} is given by:
\begin{equation}\label{eq cp1}
\mathbf{h}(\bu; f(\mathbf{p}), \mathbf{b}) = \sigma(f(\mathbf{p})\bu+\mathbf{b}).
\end{equation}

An easy way to incorporate the formulation into existing neural network models is by making $f$ a single-layer MLP, the conditionally parameterized dense (CP-Dense) layer can be represented by:
\begin{equation}\label{eq cp}
\mh(\bu,\mathbf{p}; \mW, \mB, \mb ) = \sigma\left(\sigma\left(\left<\mW,\mathbf{p}\right>+\mB\right)\bu+\mb\right).
\end{equation}

It should be noted that this would bring a change in the dimensions of weights and biases, which become $\mathbf{W}\in\re^{(n_h\times n_u)\times n_p}$, $\mB\in\re^{n_h\times n_u}$. When the layer width is kept the same, the total number of trainable parameters increases linearly with the parameter size $n_p$. In applications, $\mathbf{p}$ is not limited to an additionally-introduced parameter. When simply taking $\mathbf{u}$ as the parameter for itself, the quadratic terms will be introduced. High-order terms, which are prevalent in physical systems, can be easily modeled using multiple such layers. In Appendix~\ref{sec appendix b}, we demonstrate how certain discretized PDE terms can be fitted exactly with simple conditionally parameterized layers.

\subsection{CP-GNet for mesh-based modeling of physical systems}\label{sec cpnet}
{\bf Graph representation of discretized systems:}
Consider a physical system governed by a set PDEs for a time-variant vector of variables $\mathbf{q}(t)$. Using the popular finite volume discretization, the computational domain is divided into contiguous small cells, indexed by $i$. The discretized form of equation can be written as:
\begin{equation}
    \frac{d {\mathbf {q} }_{i}(t) }{dt} = \frac{1}{\Omega_{i}}\sum_{j \in N(i)}
 {\mathbf f} \left( \mq_i,\mq_j,\mathbf n_{ij}  \right)A_{ij}+\mathbf{s}(\mathbf{q}_i),
\end{equation}\label{eq fv}
where $\mathbf{q}_i$ is the cell-centered value of cell $i$, $\Omega_i$ is the volume (3D)/area (2D) of the cell, and $N(i)$ is the neighborhood set of cells around $i$. Between a neighboring pair of cells $i$ and $j$, $A_{ij}$ is area (3D)/length (2D) of the shared cell boundary, and $\mathbf{n}_{ij}=(\mx_i-\mx_j)/|\mx_i-\mx_j|$ is a vector between the cell center locations $\mathbf{x}_i$ and $\mathbf{x}_j$. In the explicit numerical simulation of \eqr{eq fv}, solutions are updated by computing the increment of $\Delta\mathbf{q}_i^k=\mathbf{q}_i^{k+1}-\mathbf{q}_i^k$ between discrete time steps indexed by $k$, which is determined by two terms. The \textit{flux term} $\mathbf{f}$ computes the exchange of quantity between neighboring cells, which is a complex function involving both the cell values as well as the vector between them, e.g.~\citep{toro2012godunov}. The  \textit{source term} $\mathbf{s}$ computes physics that are local to the cell, such as the reaction of chemical species.  

In our setting, the discretized system is mapped to a graph $G(V,E)$, defined by nodes $V$ of size $|V|=n_v$ connected by edges $E\subset V\times V$ of size $|E|=n_e$. Each node $\mathbf{v}_i$ is located at the corresponding cell center $\mathbf{x}_i$, and each edge $(i,j)$ corresponds to a shared boundary between the finite volume cells. Denoting the sets of mapped quantities on all nodes and edges of $G$, $\mathbf{Q}=\{\mq_i, i\in V\}, \mathbf{N}=\{\mathbf{n}_{ij}, (i,j)\in E\}$, the target is to develop a graph neural network operator $g$ that predicts the increment as $\Delta\mathbf{Q}^k=g(\mathbf{Q}^k, \mathbf{N})$.

\begin{figure}[!ht]
	\centering
	\includegraphics[width=1\columnwidth]{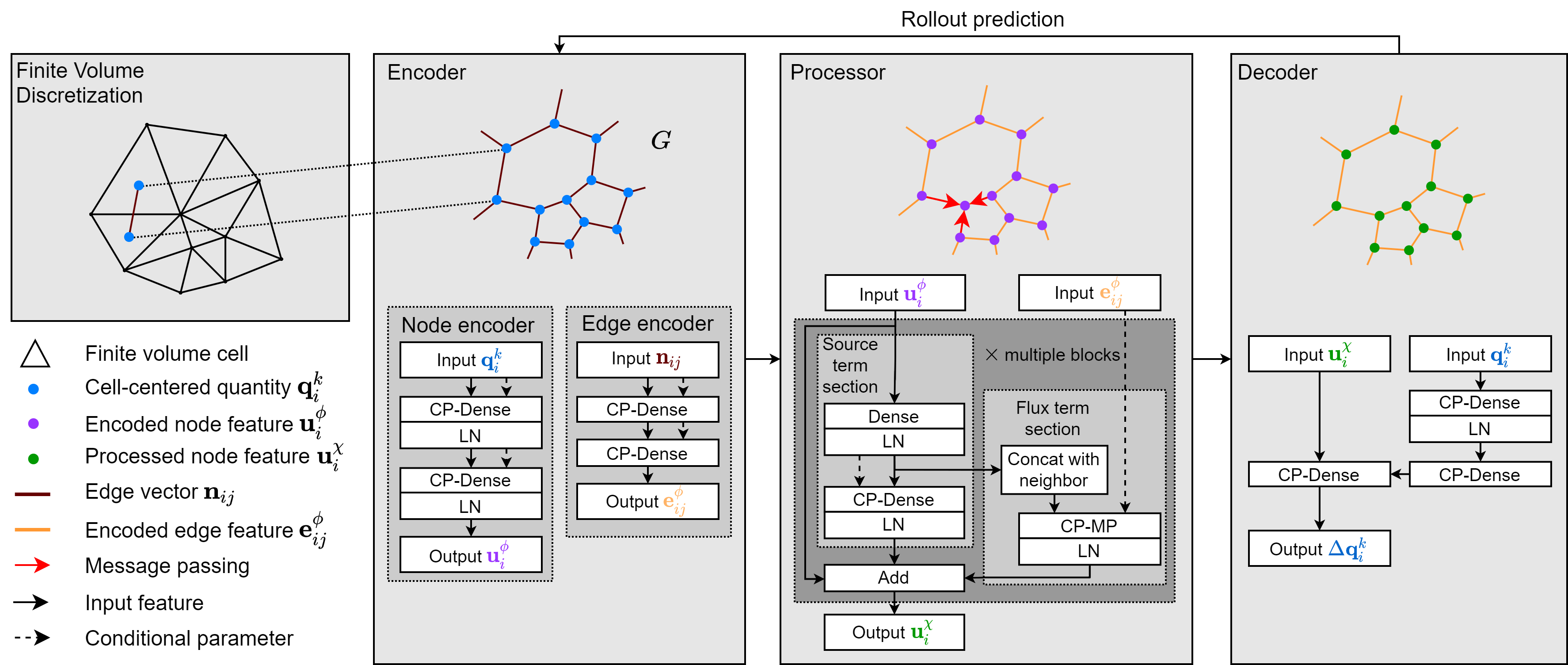}
	\caption{Schematic of CP-GNet architecture}\label{fig diagram}
\end{figure}

{\bf CP-GNet architecture:}
The  architecture for the proposed conditionally parameterized graph neural network, CP-GNet, can be written in a encoder-processor-decoder form. A schematic is provided in Fig.~\ref{fig diagram}. For clarity of different variables in the description of the network, we use $\mathbf{u}_i$ for the latent variables on node $i$ to distinguish from the physical variables $\mathbf{q}_i$, and use $\mathbf{e}_{ij}$ for latent variables on edge $(i,j)$ to distinguish from the vector $\mathbf{n}_{ij}$.

{\bf Encoder:}
Numerical solution of PDEs (e.g., the compressible Navier--Stokes equations) requires the processing of arbitrarily complex interactions between mesh elements. While large MLP architectures can represent this complexity, the data requirements to reliably train such networks might be large. In contrast, our proposed encoder takes two CP-Dense layers, taking the output from the previous layer as both the input and the conditional parameter. Through the encoder, high-order interactions can be easily extracted, allowing a degree of extrapolation by virtue of linearity. The CP-GNet uses two separate, but similarly constructed encoders to process the input node features $\mathbf{q}_i^k$ and edge features $\mathbf{n}_{ij}$, respectively. 

{\bf Processor:}
The flux term $f$ in \eqr{eq fv} can be effectively approximated by CP message-passing (CP-MP) between adjacent nodes on a graph. The source term $s$, on the other hand, can be modeled by CP-Dense layers. In CP-GNet the processor consists of multiple identical blocks with independent weights. Residual connections are added between the blocks. As shown in Fig.~\ref{fig diagram}, each block includes a CP-MP based section and a CP-Dense based section to address the two types of terms. Modified from the Edge Conditioned Convolution (ECC)~\citep{simonovsky2017dynamic}, the CP-MP computation is formulated as:
\begin{equation}\label{eq cpmp}
    \mW_{ij} = \sigma\left(\left<\mW, \me^{\phi}_{ij}\right>+\mB\right),\ \ ; \ \
    \mathbf{h}_i = \sum_{j\in N(i)}w_{ij}\sigma\left( \left<\mW_{ij},\left[ \bu_i; \bu_j\right]\right> \right),\\
\end{equation}
where $\me^{\phi}_{ij}$ is the output from the edge encoder, $\mathbf{u}_i$ and $\mathbf{u}_j$ are the latent node features from the previous layer, $\mathbf{h}_i$ is the nodal latent output, and $w_{ij}=A_{ij}/\Omega_i$ is the flux weight from \eqr{eq fv}.

{\bf Decoder:}
It is common in a PDE solver to use a Jacobian matrix $\mathbf{J}=\partial \bu/\partial\mathbf{h}$ to transform  the variable increments $\Delta\bu=\mathbf{J}\Delta\mathbf{h}$. The decoder in the GP-GNet serves a similar purpose -- to convert hidden variables to the output on the physical space. Similar to the encoder, the decoder consists of three conditionally parameterized dense layers. The first two layers can actually be viewed as a dedicated ``encoder'' that is similar to the initial node encoder, taking $\mathbf{q}_i^k$ as the input, but with independent weights. The purpose of this ``encoder'' is to extract a final conditional parameter, which is used in the third CP-Dense layer in the decoder to determine the weights for the output node feature $\mathbf{u}_i^{\chi}$ from the processor. The third decoder layer is also the final layer of the model, which outputs $\Delta\mathbf{q}_i^k$ (with proper scaling). Except for the edge encoder and the last two layers in the decoder, all dense, CP-Dense, CP-MP layers are appended with LayerNormalization (LN) layers. 

{\bf Treatments for boundaries:}
The computational domain of a practical problem includes multiple types of boundaries, e.g. the case in Sec.~\ref{sec deepblue}. In classic PDE solvers, they are treated with different boundary conditions, which define explicit formulations to compute relationships of the domain with the external world. However, these conditions and formulations are only defined for the physical quantities, thus cannot be easily transferred for latent variables, especially when multiple message-passing/convolution steps are used. To enable the GP-GNet to model different types of boundaries efficiently, special treatments are necessary.For boundaries with known inputs, such as the inlet and the outlet, their values are directly input to the corresponding nodes at every time step. For the boundaries imposing certain constraints, instead of a given physical value, such as Neumann and  symmetry boundaries, \textit{ghost edges} are introduced. For a cell $i$ with a face lying on a boundary, we introduce a ghost edge vector $\mathbf{n}_{ig}$, that points from the corresponding node $i$ to the center of the boundary face. Ghost edges are processed together with the normal edges in the edge encoder. However, the CP-MP layer in the processor of the CP-GNet is slightly modified. More specifically, the concatenation $[\mathbf{u}_i;\mathbf{u}_j]$ in \eqr{eq cpmp} is replaced with only $\mathbf{u}_i$. And for each type of boundary, the weights for the CP-MP layer are trained independently, to let the model learn different types of boundary condition for the latent variables. The effectiveness of this treatment is shown in Sec.~\ref{sec deepblue} and further discussed in Appendix.~\ref{appendix additional rocket}.

\vspace{-0.2cm}
\section{Related Work}\label{sec related works}
\vspace{-0.2cm}

There have been successful attempts towards making networks directly parametric to certain features, such as connectivity patterns~\citep{stanley2009hypercube}, layer embedding~\citep{ha2016hypernetworks}, mean image features~\citep{yang2019condconv}.  The Conditionally Parameterized Convolution (CondConv) model~\cite{yang2019condconv}, makes convolution kernel weights as a linear combination of functions of the input features, and achieves an efficient expansion of the network capacity. The Hypernetwork~\cite{ha2016hypernetworks} uses a single network that takes layer embeddings, e.g., layer index, to generate the weights for different layers of the main network, and reduced the total number of trainable weights. A popular framework to perform convolution on graphs is the message passing neural network (MPNN)~\cite{gilmer2017neural}, which treats graph convolutions as messages passed between nodes through edges.  In this approach, the node features and edge act on intermediate variables and the output is expressed as a linear combination through concatenation. This can fail when the impact of node features rely on the edge features in a non-linear fashion. To address this, Edge Conditioned Convolution~\cite{simonovsky2017dynamic} (ECC) makes the weights for node features dependent on edge features.
After the modification for conditional parametrization, ECC was shown to achieve excellent performance on irregular point cloud data. In comparison, our method extend the choice of parameters to physical quantities, hidden inputs themselves, as well as discretization information.

Multiple architectures in the family of GNNs have shown successes in processing irregular, non-Euclidean features. Applications include cloud classification~\citep{landrieu2018large, wang2019dynamic}, action recognition~\citep{yan2018spatial} and control~\citep{sanchez2018graph}, traffic forecasting~\citep{yu2017spatio, zhang2018gaan}, quantum chemistry~\citep{gilmer2017neural}. Attempts on using GNNs in scientific computation are relatively limited and are mostly focusing on particle-based methods~\citep{li2018learning, sanchez2020learning}. Recently, pioneering work has demonstrated the potential of using GNNs for mesh-based scientific  computation. CFD-GCN~\cite{belbute2020combining} coupled a GNN with an existing PDE solver to perform hybrid-fidelity prediction and achieved higher efficiency than traditional high-fidelity solvers. MeshGraphNets~\citep{pfaff2020learning} extends the encoder-processor-decoder structure from Graph Network-based Simulators (GNS)~\citep{sanchez2020learning}, and demonstrated impressive performance on mesh-based simulations for a wide range of physical systems. Compared to these approaches, our method with CP models the high-order terms and irregular discretizations more effectively. Appendix~\ref{appendix comparison} compares our method with the MeshGraphNets on flow simulation tasks.
\vspace{-0.1cm}
\section{Numerical Tests}\label{sec result}
\vspace{-0.1cm}
We applied conditional parametrization to network architectures for three distinct, but important tasks in scientific computing. The first two tasks are on uniform Euclidean grids, and the discretization information is directly included in the conditional parameters such as the differential terms and the local Reynolds number. Comparisons between appropriate baseline models and their CP modifications are performed. The third task uses a irregular mesh with complex boundaries and is conducted with the CP-GNet model we proposed. The non-CP modification, which to our knowledge fall into a similar architecture to that for the MeshGraphNets~\citep{pfaff2020learning} is used as the baseline. Appendix ~\ref{appendix dataset} provides more details on the studied system and the generation of data; ~\ref{appendix network} provides details on network training; ~\ref{appendix additional} provides additional results and analysis. An additional test for the flow over a cylinder is performed in the comparison against the MeshGraphNets in Appendix~\ref{appendix comparison}.

\subsection{Discovery and solution of coarse-grained models}\label{sec closure}
In many practical problems, high fidelity simulations are not affordable. Instead, computations are performed using coarse-grained models, e.g. the Large Eddy Simulation~\cite{moin2002advances}. In such models, the small-scale physics are unresolved, and are approximated using additional \textit{closure} terms in the PDEs, the development of which constitutes an important area of research. In fact, even for the seemingly simple (yet richly non-linear) equation presented below, a perfect closure model is unknown. In this work, we demonstrate how CP models can be used to develop a closure model for the coarse-grained 1D viscous Burgers equation that is often used in the study of shock formation, traffic flows, and turbulent interactions, etc. For the unknown spatio-temporal field $u(x,t)$ on a spatially periodic domain  $x\in[0,L]$, the original equation is given by:
\begin{equation}\label{eq 1d burgers}
    \frac{\partial u}{\partial t} +u\frac{\partial u}{\partial x}- \nu\frac{\partial^2 u}{\partial x^2}=0,
\end{equation}
where $\nu$ is a diffusion coefficient and $u(x,0)$ is a random initial condition (See Appendix~\ref{appendix dataset closure}).

When this equation is solved on a finely discretized mesh, the dynamics can be regarded as fully resolved. However, if a solution is attempted on a coarse mesh with \eqr{eq 1d burgers} without any additional treatments, the solution becomes inaccurate and numerically unstable, thus a closure operator $\mathcal{C}(\cdot)$ is needed. Representing the quantity on the lower resolution mesh by $\bar u$, the ``closed'' equation is:
\begin{equation}\label{eq closure}
    \frac{\partial \bar u}{\partial t}+ \bar u\frac{\partial \bar u}{\partial x} - \nu\frac{\partial^2 \bar u}{\partial x^2} + \mathcal{C}=0.
\end{equation}

\begin{figure}[!ht]
	\centering
\hspace*{\fill}
	\subfloat[CNN]{
    	\includegraphics[height=0.105\columnwidth]{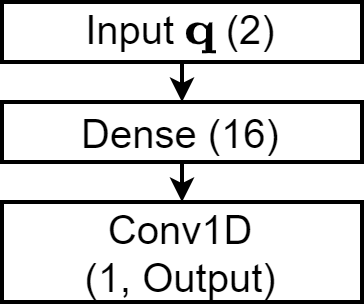}
    }
\hspace*{\fill}
	\subfloat[CP-CNN]{
    	\includegraphics[height=0.105\columnwidth]{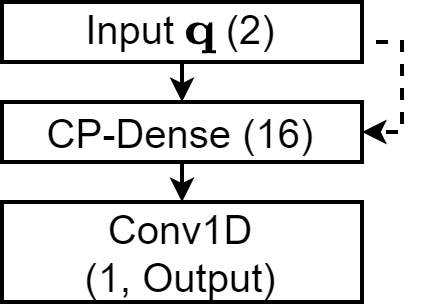}
    }
\hspace*{\fill}
	\subfloat[DDP]{
    	\includegraphics[height=0.16\columnwidth]{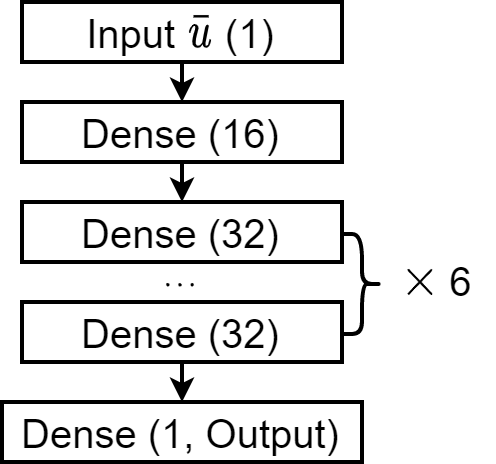}
    }
\hspace*{\fill}
	\subfloat[CP-DDP]{
    	\includegraphics[height=0.16\columnwidth]{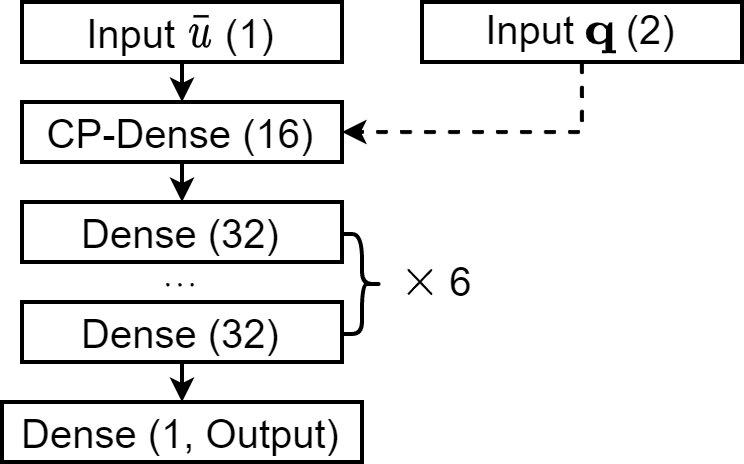}
    }
\hspace*{\fill}
	\caption{Closure modeling network architectures. Solid arrow: input feature; dashed arrow: condition parameter; numbers: layer width.}\label{fig closure arch}
	
\end{figure}

In this experiment, two baseline models for $\mathcal{C}$ and their CP developments are compared. The first model is 2-layer CNN with a dense layer with ReLU activation, followed by a 1D convolution layer. This model assumes the closure term to be a function of convection term $\bar u\frac{\partial \bar u}{\partial x}$ and the diffusion term $\nu\frac{\partial^2 \bar u}{\partial x^2}$, and takes their concatenation $\mathbf{q}=[\bar u\frac{\partial \bar u}{\partial x}, \nu\frac{\partial^2 \bar u}{\partial x^2}]$ as the input. Its CP variant, CP-CNN, replaces the first layer with a CP-Dense layer that takes $\mathbf{q}$ as the parameter for its own weights. The second baseline model is a reference Data-Driven Parameterization (DDP) model~\citep{subel2021data}. The model takes $\mathcal{C}$ as a function of the filtered variable $\bar{u}$, which is modeled by an 8-layer MLP with swish activation. Similarly, the CP variant, CP-DDP replaces the first layer with a CP-Dense layer that takes $\mathbf{q}$ as the parameter for the weights for $\bar{u}$. The network architectures are presented in Fig.~\ref{fig closure arch}.  




Two sets of data are used. The high resolution runs are solved with \eqr{eq 1d burgers} from two different initial conditions (ICs) on a shared 2048-grid-node mesh. The low resolution solutions are obtained by applying a box-filter to each step of the high resolution solutions onto a 32-grid-node mesh. The ground truth for $\mathcal{C}$ is then computed based on the low resolution data. Each set of data consists of 267 time steps, spanning a period of 2 s. The first 0.2 s of data for one IC is used for training.

\begin{figure}[!ht]
	\centering
	\subfloat[Training IC]{
	\includegraphics[width=0.49\columnwidth]{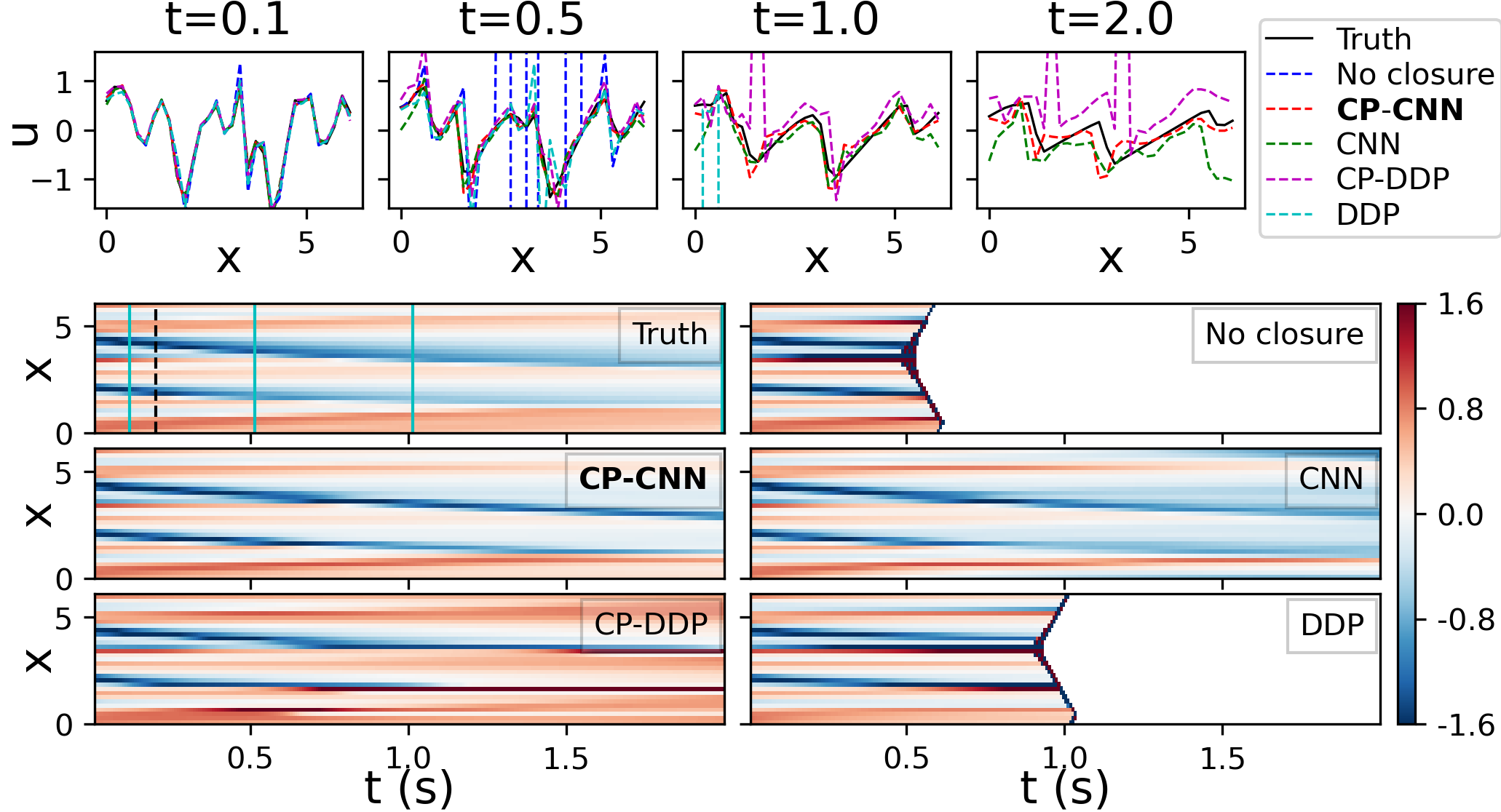}
    }
	\subfloat[Testing IC]{
	\includegraphics[width=0.49\columnwidth]{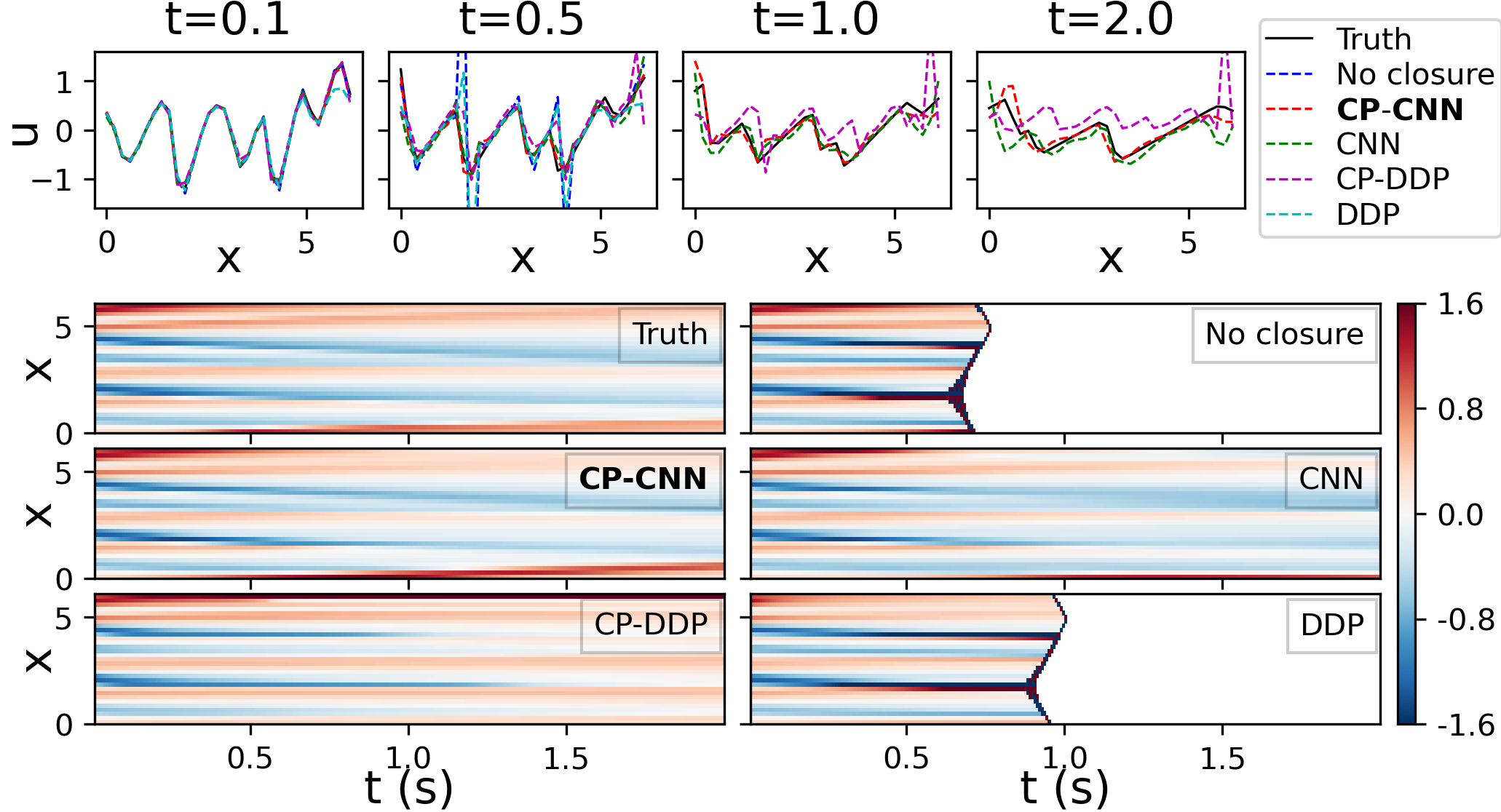}
    }
    \hspace*{\fill}
	\caption{Closure modeling results. The first $t\leq0.2$~s for the left case is used for training, marked by the black dashed line in the first contour. The $x$-$t$ contours show the evolution of $\bar{u}$. \textbf{The reference DDP model solution grows into infinity, shown as white areas in the contour}. The gaps between models are more visible in the spatial profiles at time steps marked by the cyan lines.}\label{fig closure}
\end{figure}

Online testing computations are then carried out from the filtered, low resolution ICs using \eqr{eq closure}, with $\mathcal{C}$ computed based on the online solution at every time step. $x$-$t$ contours are present in Fig.~\ref{fig closure} to compare the evolution of $\bar{u}$. Spatial profiles are also plotted at a few steps to provide more details. Despite a small time step (CFL number$< 0.5$, without any closure term, the computation is numerically unstable and the error grows unbounded. The baseline CNN model is able to keep the solution stable within the period studied, and the CP-CNN improves the accuracy noticeably. The baseline DDP model is only able to postpone the ``blow-up'' to slightly later. The solution with CP-DDP closure is bounded throughout the period. The improevments are also valid for both the unseen IC. The Mean Absolute Error (MAE) for $\bar{u}$ is provided in Table.~\ref{table closure}. 

\begin{table}
\small
\caption{Closure model MAE. $\bar{u}$ Avg.: averaged over all steps for online prediction for $\bar{u}$; $\bar{u}$ final: for the final step of online prediction; Inf.: Unbounded cases.}\label{table closure}
\centering
\begin{tabular}{c|cc|cc}
\hline
\multirow{2}{*}{} & \multicolumn{2}{c|}{Training IC} & \multicolumn{2}{c}{Testing IC}                               \\ \cline{2-5} 
                  & $\bar{u}$ Avg. & $\bar{u}$ final & $\bar{u}$ Avg. & $\bar{u}$ final\\ \hline
CNN &0.23 &0.41&0.16 &0.23\\
CP-CNN &\textbf{0.15} &\textbf{0.21} &\textbf{0.09} &\textbf{0.13} \\
DDP &Inf. &Inf. &Inf. &Inf.\\
CP-DDP &0.42 &0.89 &0.3 &0.41\\ 
\hline
\end{tabular}
\end{table}


\subsection{Super-resolution of chaotic flows}\label{sec superres}
\begin{figure}[H]
    \begin{minipage}[c]{0.5\textwidth}
    \centering
    \includegraphics[trim=42 5 37 10,clip,height=0.53\linewidth]{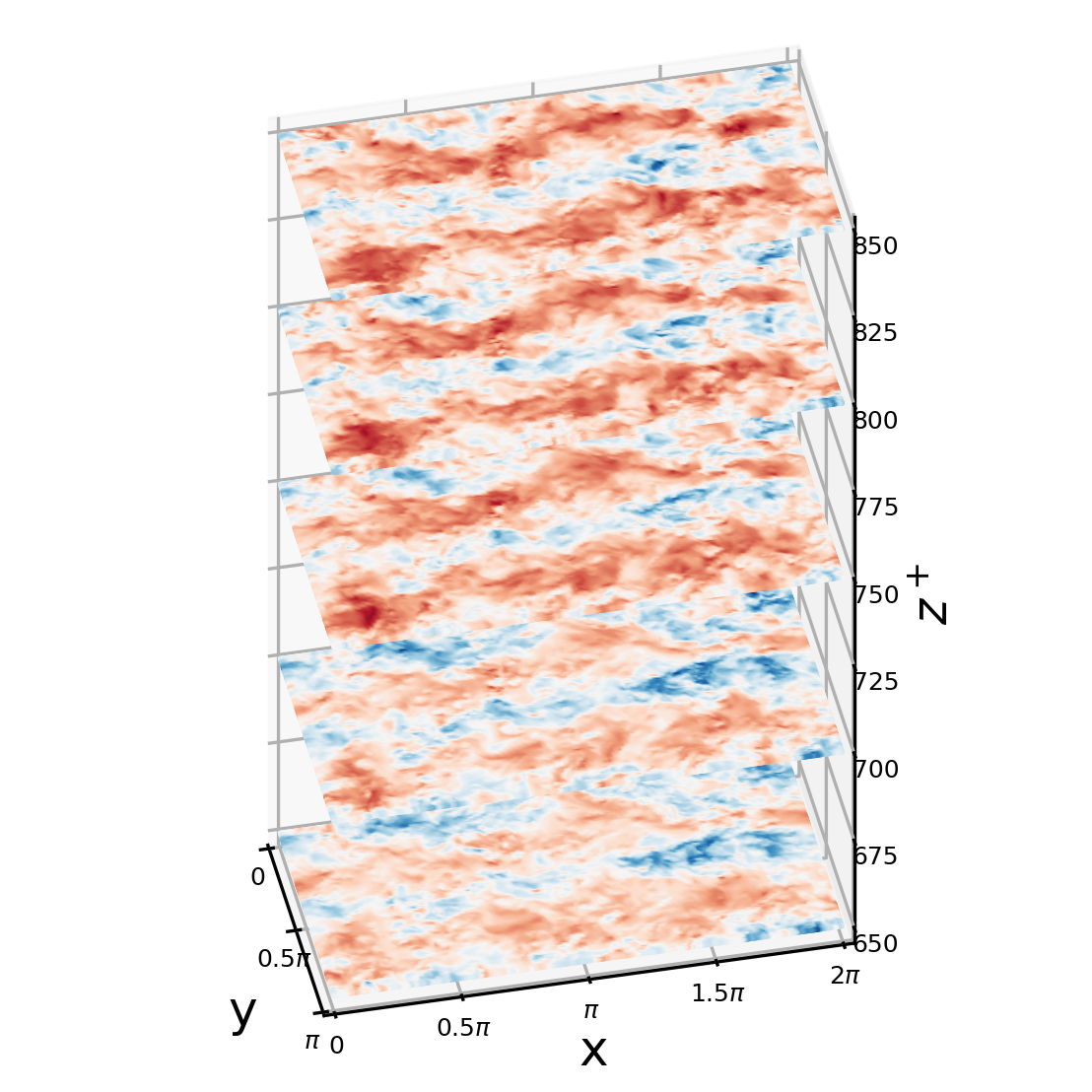}
    \includegraphics[height=0.53\linewidth]{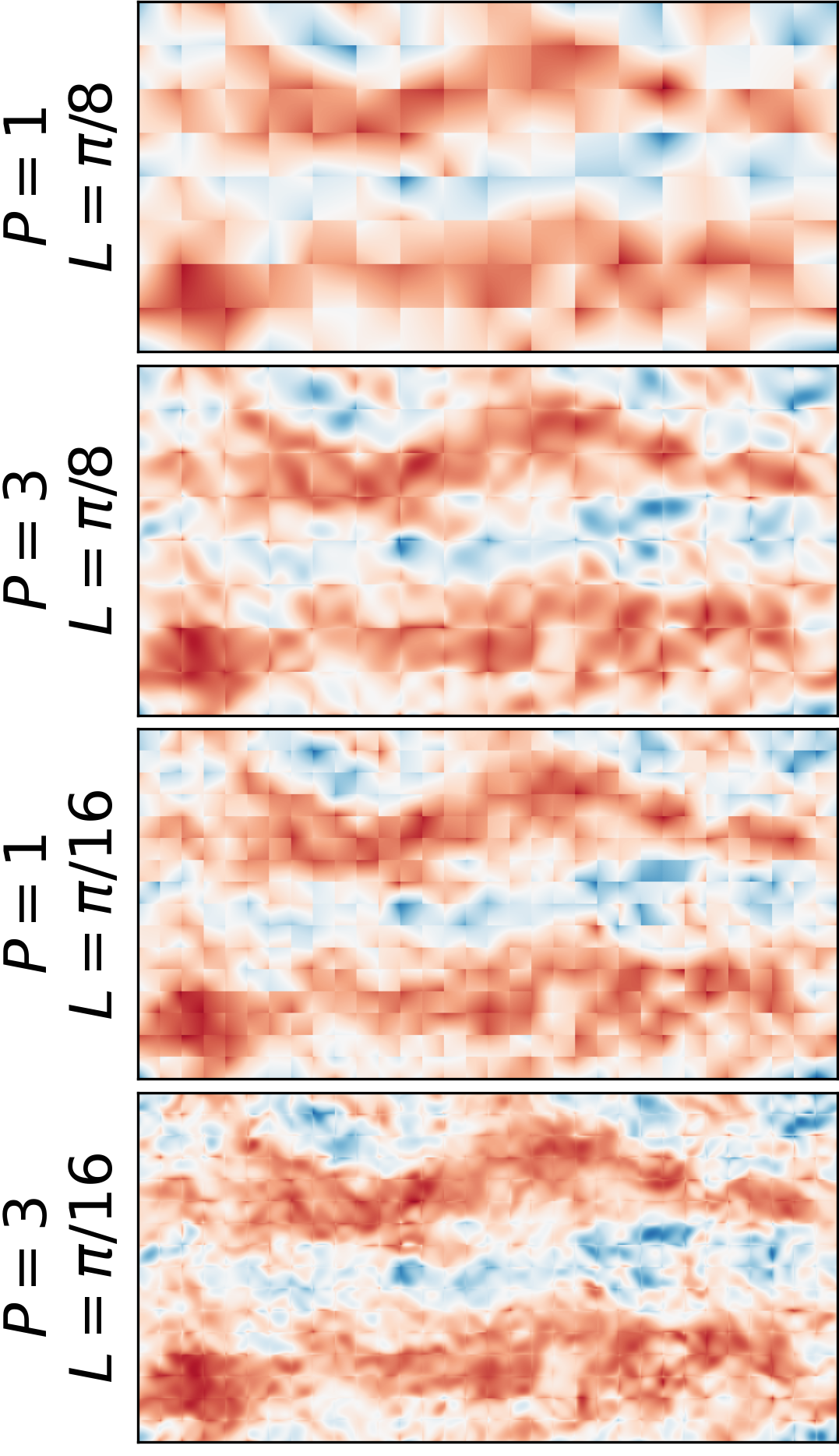}
    \caption{Snapshots for super-resolution}\label{fig superres truth}
    \end{minipage}
    \hspace*{\fill}
    \begin{minipage}[c]{0.48\textwidth}
    \centering
    \hspace*{\fill}
        \subfloat[MLP]{
        \includegraphics[height=0.363\linewidth]{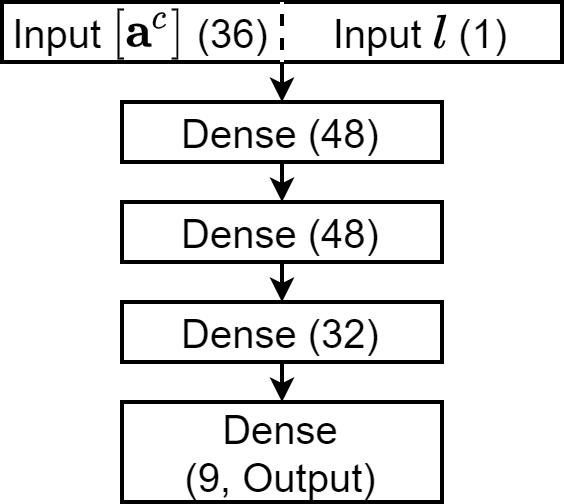}
        }
    \hspace*{\fill}
        \subfloat[CP-MLP]{
        \includegraphics[height=0.363\linewidth]{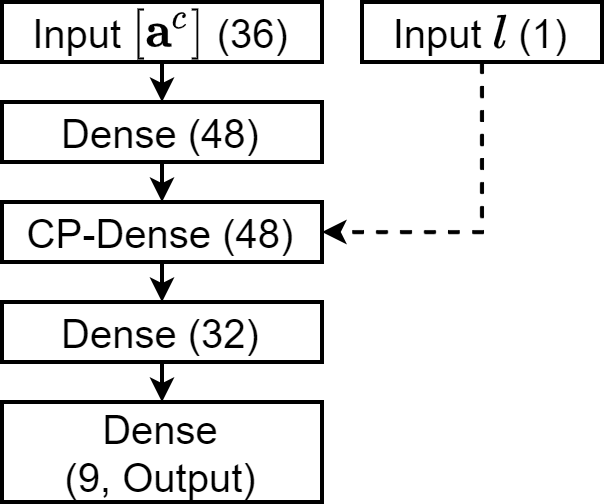}
        }
    \hspace*{\fill}
    \caption{Super-resolution network architectures Solid arrow: input feature; dashed arrow: condition parameter; numbers: layer width.}\label{fig superres arch}
    \end{minipage}
    \hspace*{\fill}
\end{figure}

In this experiment, we perform enrichment of low-resolution snapshots of turbulent flow fields. In an enrichment/super-resolution process, one inputs a low-resolution snapshot of the solution, and seeks a snapshot with better resolution. One way to achieve different resolutions on a given mesh is to use Discontinuous Galerkin (DG) projection~\cite{cockburn2012discontinuous}. In this method, the solution within a mesh element $i$ is represented by coefficients $\mathbf{a}_i$ for a set of polynomial bases, of which the size is determined by the polynomial order $P$. The final resolution of the solution is jointly determined by $P$ and the element width $L$. 
More specifically, wall-parallel snapshots from the solution of a turbulent channel flow~\citep{del2004scaling} is studied, and the task is to recover high-order ($P=3$) DG coefficients $\mathbf{a}^h_i\inr{9}$ for the $x$-velocity from lower-order ($P=1$) ones $\mathbf{a}^l_i\inr{4}$. 5 snapshots are generated in total at different normalized wall-normal heights $z^+\in\{650, 700, 750, 800, 850\}$, as illustrated in Fig.~\ref{fig superres truth}. Each snapshot spans an area of $X\times Y=2\pi\times\pi$, and is projected onto a shared set of uniform meshes with 6 different widths $L\in\{\pi/4, \pi/8, \pi/12, \pi/16, \pi/24, \pi/32\}$, for the two studied polynomial orders $P\in\{1,3\}$. Thus, for each $z^+$, 12 sets of data, each for one combination of $L$ and $P$, are provided. Fig.~\ref{fig superres truth} shows a few example contours at different combinations for $z^+=800$. The data for $z^+\in{700,800}$ is used for training. It should be noted that the coefficients are computed independently for each mesh element, thus the total number of training points is a few thousand, instead of 24 (which should be multiplied by the number of elements). More details on the data generation are provided in Appendix~\ref{appendix dataset super}.

In this task, the baseline model is from the compact super-resolution model by \citet{pradhan2021variational}. It takes $\mathbf{a}^h_i$ as a function of two inputs. The first input is a concatenation of normalized low-order basis coefficients for $i$ and its neighbors $N(i)$:
\begin{equation}
    [\mathbf{a}^c]_i=[\{\mathbf{a}^c_j-\bar{\mathbf{a}}^c;j\in N(i)\cup{i}\}]/u^{\mathrm{RMS}}_i,
\end{equation}
where $[\{\cdot\}]$ denotes the concatenation of all elements in a set, and $\bar{\mathbf{a}}^c$ is the mean of the set. In our case, we include all immediate neighbors, including corner ones in $N(i)$, thus $[\mathbf{a}^c]_i\inr{36}$. The second input to the model is an indicator $l_i=\log(Re^L_i)$ for the loss of information in the low-order projection process. $Re^L_i=\frac{u^{\mathrm{RMS}}_iL}{\nu}$ is the local Reynolds number. The indicator reflects that the loss is a function of the kinetic energy, measured by $u^{\mathrm{RMS}}_i$, mesh resolution $L$, and fluid viscosity $\nu$. Because $Re^L_i$ can vary by orders of magnitude across elements, log scaling is used. The two inputs are first concatenated and then processed in a 4-layer MLP in the baseline model. In contrast, the conditionally parameterized model CP-MLP processes only the first input $[\mathbf{a}^c]_i$ in the dense layers. The second dense layer is replaced by a CP-Dense layer, where the second input $l_i$ is instead taken as a conditional parameter for the weights for the latent output of the first layer. A comparison of the model architectures are provided in Fig.~\ref{fig superres arch}.

\begin{figure}
	\centering
	\includegraphics[width=1\columnwidth]{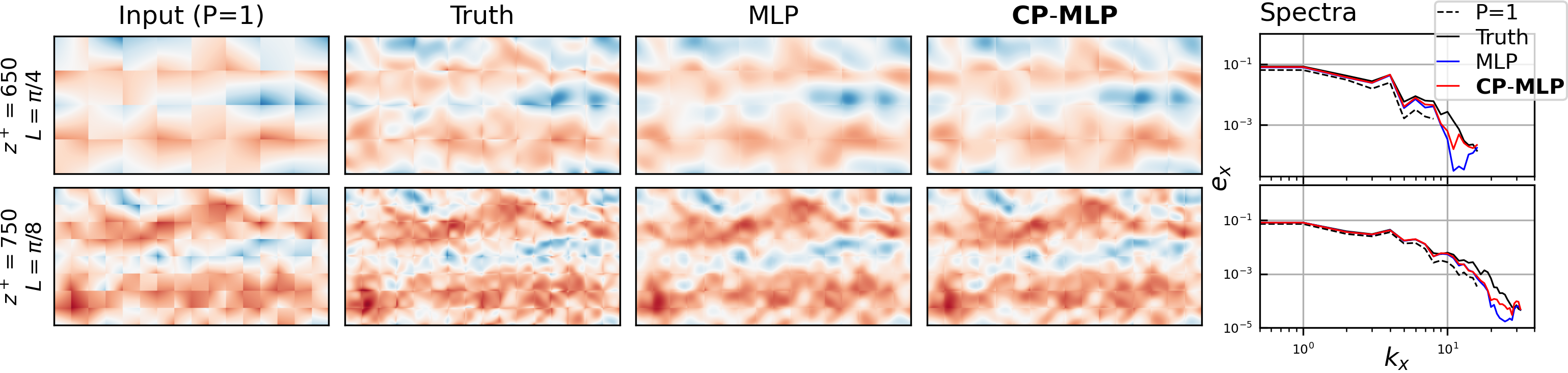}
	\caption{Super-resolved flow field and stream-wise energy spectra $e_x$ for example test cases $(z^+=650, L=\pi/4)$ and $(z^+=750, L=\pi/8)$. \textbf{The CP-MLP shows finer details on the edge of elements (adjacent squares)}, showing a better prediction of high-order coefficients. The observation is proved by a richer high $k_x$ energy spectra in the right plot.} \label{fig superres result}
\end{figure}

Results for two sample testing cases, $(z^+=650, L=\pi/4)$ and $(z^+=750, L=\pi/8)$ are shown in Fig.~\ref{fig superres result}. It can be observed that the CP-MLP is able to reconstruct more small scale structures compared with the MLP. The performance can be qualified by the stream-wise and span-wise energy spectra, $e_x$ and $e_y$ (see Appendix~\ref{appendix additional super} for definitions). $e_x$ for different stream-wise wave numbers $k_x$ is shown in Fig.~\ref{fig superres result}. It can be observed that for high-order projection or super-resolution, the high-wave-number spectra is much richer than that for the low-order projection. The CP-MLP plots follow the truth noticeably better than the MLP baseline, which confirms our observation from the contours. Absolute error in the integrals of energy spectra, $E_x=\int_{k_x}e_xdk_x$ and $E_y=\int_{k_y}e_ydk_y$ are computed for the 24 training and 36 testing sets and summarized in Table~\ref{table superres}. Both training and testing errors are reduced significantly when CP is applied.

\begin{table}
\small
\caption{Average and maximum absolute errors in the integral of super-resolved energy spectra.}\label{table superres}
\centering
\begin{tabular}{c|cccc|cccc}
\hline
\multirow{2}{*}{} & \multicolumn{4}{c|}{Training}                             & \multicolumn{4}{c}{Testing}                               \\ \cline{2-9} 
                  & $E_x$ Avg. & $E_x$ Max. & $E_y$ Avg. & $E_y$ Max. & $E_x$ Avg. & $E_x$ Max. & $E_y$ Avg. & $E_y$ Max. \\ \hline
MLP & 0.0145 & 0.0391 & 0.0272 & 0.0609 & 0.0098 & 0.0364 & 0.0184 & 0.0675 \\
CP-MLP & \textbf{0.0120} & \textbf{0.0328} & \textbf{0.0217} & \textbf{0.0429} & \textbf{0.0081} & \textbf{0.0260} & \textbf{0.0158} &\textbf{0.0519}\\ \hline
\end{tabular}
\end{table}


\subsection{Simulation of reacting flows in a rocket  engine injector}\label{sec deepblue}
\begin{figure}[!ht]
    \begin{minipage}[c]{0.47\textwidth}
	\centering
	\includegraphics[width=0.8\columnwidth]{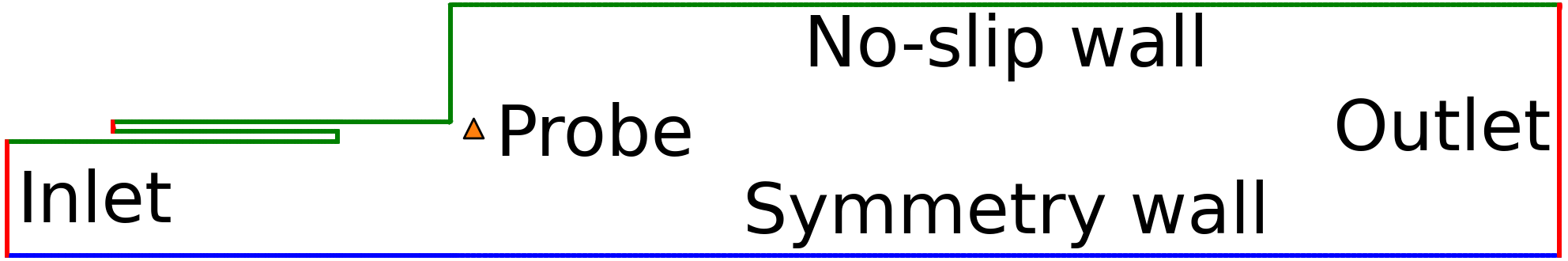}
	\vskip0.5em
	\includegraphics[width=0.9\columnwidth]{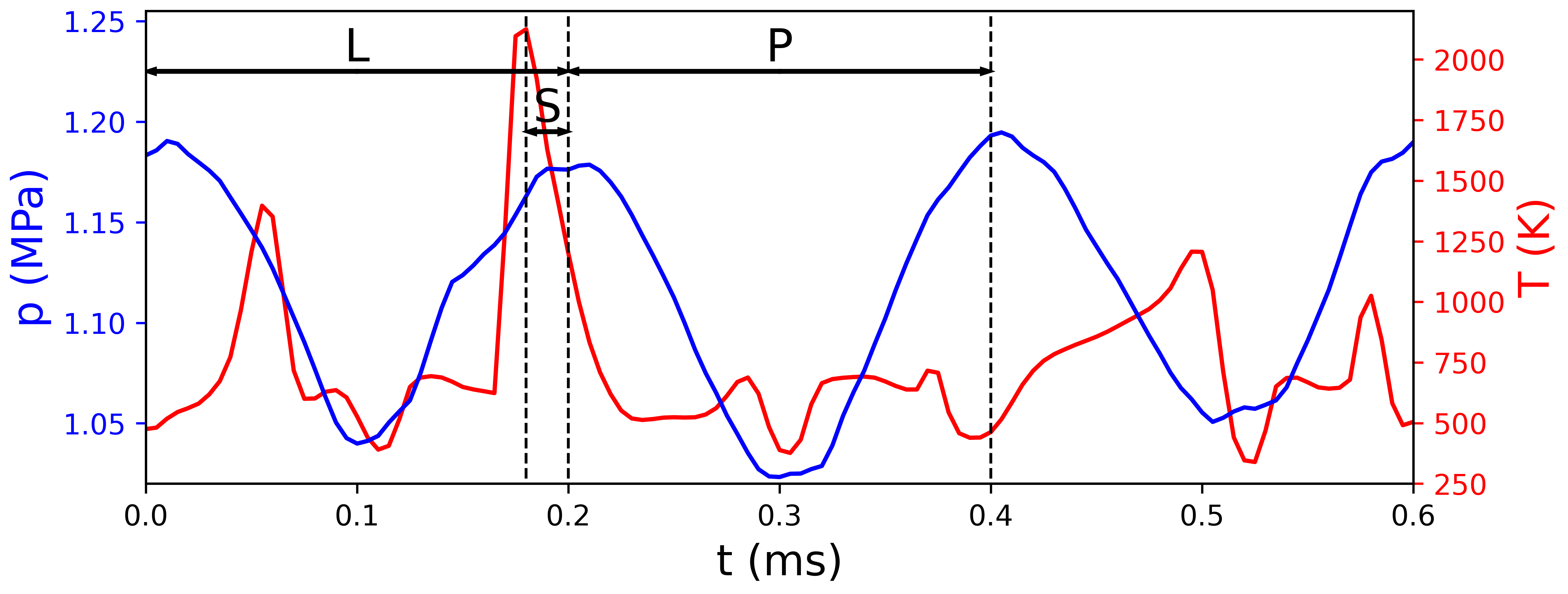}
	\caption{Injector outline and probed response for $p$ and $T$. Orange marker: probe location. L/S: long (0.2 s)/short (0.02 s) training period (0.2 s); P: prediction period (0.2 s).}\label{fig perturb}
	\end{minipage}
    \hspace*{\fill}
    \begin{minipage}[c]{0.48\textwidth}
	\centering
	\includegraphics[width=1\columnwidth]{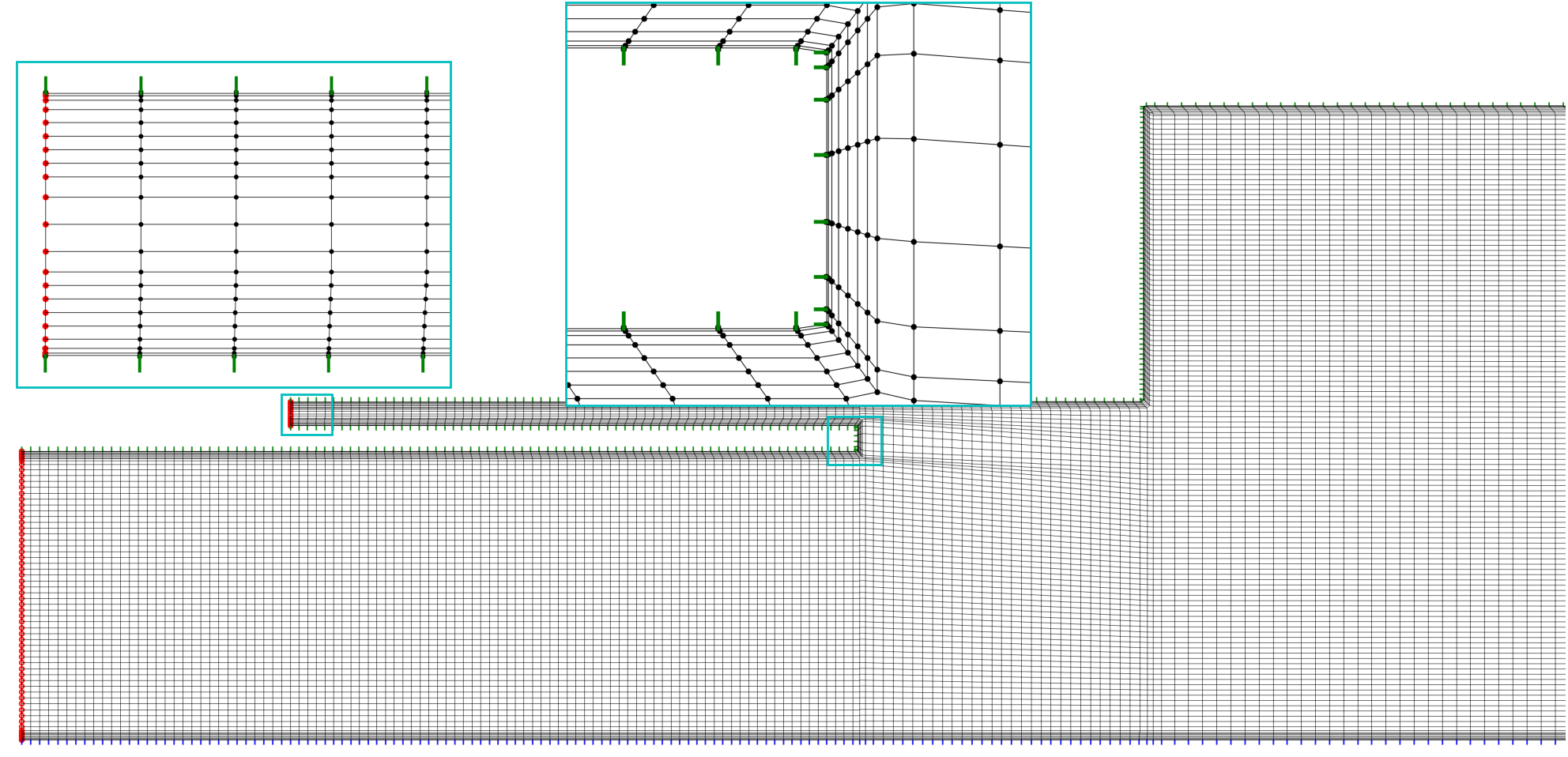}
	\caption{Graph details. Black dots: standard nodes; black lines: standard edges; red dots: inlet/outlet nodes; green/blue lines: two groups of ghost edges (extruded for visualization).}\label{fig mesh}
    \end{minipage}
\end{figure}

We use 
 a highly complex public dataset~\cite{huang2019investigations} as a model of combustion processes in a rocket engine injector~\cite{huang2021model}. The dataset includes solutions on a 2D finite-volume mesh with  308184 unknowns {\em at every time instant}. This  includes eight variables at each discretized cell: $\mathbf{q}=[p, u, v, T, Y_{\text{CH4}}, Y_{\text{O2}}, Y_{\text{H2O}}, Y_{\text{CO4}}]^T,$ where $p$ is the pressure, $u$ and $v$ are the $x$ and $y$ velocity components, $T$ is the temperature and \{$Y_{\text{CH4}}, Y_{\text{O2}}, Y_{\text{H2O}}, Y_{\text{CO4}}$\} are the mass fractions for the chemical species involved in the combustion process. The injector is outlined in Fig.~\ref{fig perturb}, where the oxidizer (O2 diluted in H2O vapor) and fuel (CH4) are injected from two inlets, respectively, into a tube-like combustion chamber in which they mix and react. The products are exhausted through an outlet. A probe monitor is placed inside the physics-intensive area, which is also marked in the figure. The strong instabilities in the simulation is triggered by a strong 2000 Hz pressure perturbation at the outlet. Fig.~\ref{fig perturb} shows the responses for $p$ and $T$ at the probe. It should be noted that, although the pressure perturbation at the outlet is periodic, the upstream behavior is affected by complex coupled physics and is not as periodic, especially for other variables such as $T$. Fig.~\ref{fig mesh} shows the graph generated following the method in Sec.~\ref{sec cpnet}, where special nodes and edges, as well as irregular local structures are provided in zoomed-in views. Two groups of ghost edges are used, corresponding to two types of wall boundary conditions in the simulation: no-slip and symmetry, respectively.

In this experiment, we attempt to predict the future states of $\mathbf{q}$ using the CP-GNet introduced in Sec.~\ref{sec cpnet}. Two CP-GNets of two different depths, with a 5-block and a 10-block processor respectively, are tested. Both CP-GNets work with an encoded node feature size of 36, and encoded edge feature size of 4. The baseline model for comparison replaces all CP layers with standard dense layers of 128 units. More specifically, after the replacement, the layers taking node features as conditional parameters will retain the original input. The layers originally taking edge features as conditional parameters will take a concatenation of the original inputs and the edge features as the new input. 
The non-CP model is referred to as the GNet.  GNets, with a 10-block and a 15-block processor respectively, are studied.

\begin{figure}[!ht]
	\centering
	\includegraphics[width=1\columnwidth]{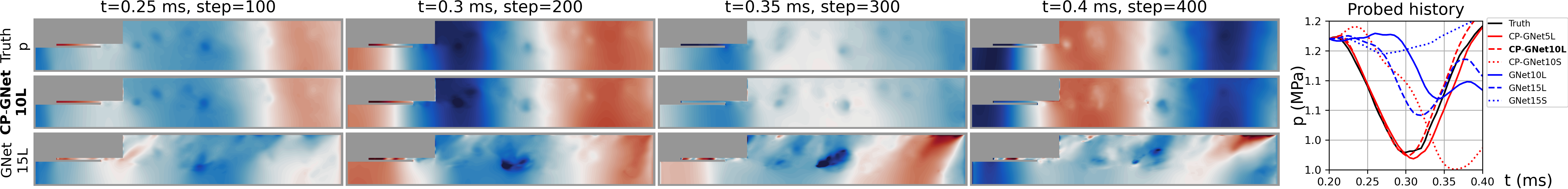}
	\includegraphics[width=1\columnwidth]{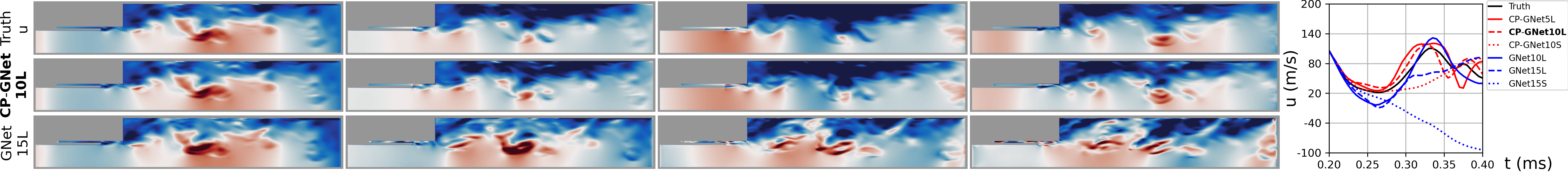}
	\includegraphics[width=1\columnwidth]{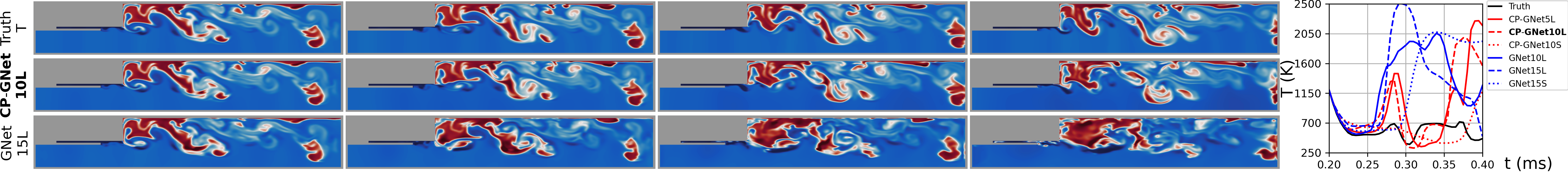}
	\includegraphics[width=1\columnwidth]{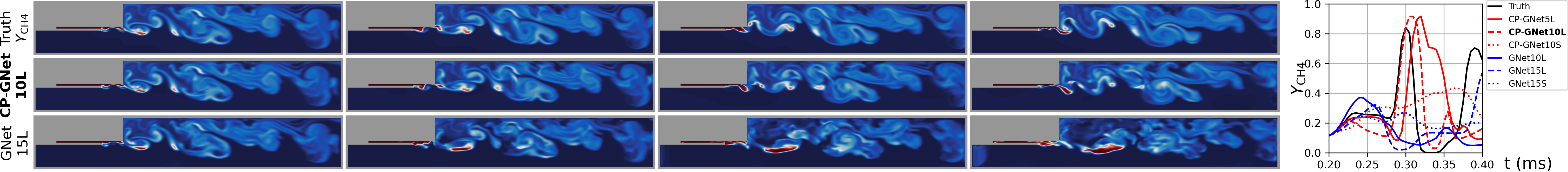}
	\caption{Predicted reacting flow. From top to bottom: pressure $p$, velocity $u$, temperature $T$, mass fraction $Y_{\mathrm{CH4}}$. \textbf{CP-GNET maintains high level of accuracy over multiple scales of mesh resolution and near complex geometry boundaries.}}\label{fig deepblue}
\end{figure}

The simulation results sampled at a time interval of $5 \times 10^{-4}$ ms are used as the ground truth. Tests  are conducted on two different lengths of training data. The long period consists of 400 steps, spanning 0.2 ms, the last 10\% of which is used as the short training period. Thus, both periods end at the same point, and rollout prediction is carried out from the end of training for another 0.2 ms. These periods are illustrated in Fig.~\ref{fig perturb}. For simplicity, we add number of processor blocks and L (long) or S (short) as suffix to the model names to distinguish them. For example, ``CP-GNet10L'' refers to the CP-GNet with 10 processor blocks trained on the long period. The predictions for 4 representative variables, $p, u, T, Y_{\text{CH4}}$, from the two deeper models, CP-GNet10L and GNet15L, are visualized in Fig.~\ref{fig deepblue} at 4 steps evenly spanned over the prediction period. The probed results are also plotted, which also covers the other models tested. It is notable that a small phase shift in the resolved structures can cause a high level of deviation in the probe measurements, and thus the flow field contours should be viewed as broader indicators of the performance.

It is  seen that the CP-GNET predicts the evolution of the reacting flow accurately over  hundreds of  prediction steps. In comparison, the non-CP model deviates quickly from the ground truth within 100 steps. Even with a smaller model (CP-GNet5L, 1.3M parameters), or a small fraction of training data (CP-GNet10S), the CP models still show comparable or even better performances compared with the largest baseline (GNet15L, 1.8M parameters). There is no significant difference in the level of error across the predicted field from our model, in spite of the vast changes in mesh density and distortion, whereas the GNets clearly suffer from more errors around the inner corners, where the mesh is the most irregular. This shows that, by combining CP with graph, discretization information can be efficiently processed. The proposed boundary treatment is also proven successful even in such a complex case with multiple types of boundaries (see Appendix~\ref{appendix additional rocket} for results without ghost edges).




\section{Summary}
This work draws inspiration from discretized numerical methods, and generalizes the idea of conditional parametrization for mesh-based models. Conditionally-parameterized networks can flexibly incorporate physical quantities as well as numerical discretization information into trainable weights, leading to efficient learning of high-order and unstructured features. Drop-in modifications are demonstrated on different architectures for several important tasks related to mesh-based modeling of physical systems. Considerable performance improvements are achieved in the numerical tests compared with the traditional counterparts. In the coarse-graining and super-resolution tasks, a small network with a simple CP-Dense layer is capable of stabilizing or improving numerical solutions. In a test of future state prediction of a rocket injector, the CP-GNet is shown to be capable of predicting the flow with complex combustion process for a few hundred steps on an irregular mesh. Although a direct CP modification will cause a linear increase in the number of parameters w.r.t. the chosen parameter, such an increase can be compensated by reducing the size of the latent vectors. Indeed, the CP-GNet is more efficient than the non-CP variant with only a fraction of the training data or with a more shallow architecture. In the appendix, we  compare the CP-GNet with the MeshGraphNet on two flow simulation tasks. Overall, the proposed architecture improves the potential for incorporating physical intuition as well as knowledge of numerical discretization. 
\section*{Acknowledgments}
J.X and K.D acknowledge support from the Air Force under the Center of Excellence grant titled {\em Multi-Fidelity Modeling of Rocket Combustor Dynamics}. A.P. is supported by NASA under the grant \#80NSSC18M0149. We thank Alvaro Sanchez-Gonzalez and Peter Battaglia for valuable advice on training noise injection for robust prediction.

\bibliographystyle{elsarticle-num-names}
\bibliography{ref.bib}

\appendix
\section{Supplemental Details}
\subsection{Data Generation}\label{appendix dataset}
\subsubsection{Closure modeling}\label{appendix dataset closure}
For the present case, the initial condition is given by:
\begin{equation}
    u(x,0) = \sum_{k=1}^{8}\sqrt{2E(k)}\sin(kx+\beta_k),
\end{equation}
where for each $k$, $\beta_k\sim\mathcal{U}(-\pi,\pi)$, and $E(k)=\max(k,5)^{-5/3}$. Other choices of parameters include domain length $L=2\pi$, viscosity $\nu=0.01$.

The 2048 mesh point high-resolution solution is generated using the Fourier-Galerkin spectral method~\cite{basdevant1986spectral} with 4th order Runge-Kutta method for time stepping. From the box-filtered initial condition, the 32-point low-resolution solution is conducted using central differencing for the spatial derivatives. This choice does not introduce additional artificial viscosity; thus, the solution without closure is naturally unstable. The high-resolution is computed at a small time-step, yet is down-sampled temporally at an interval equal to the low-resolution time step size $\Delta t$=0.0075 s. 

In this setting, $u$ can be regarded as fully resolved, thus the numerical residual $r(u)$, defined in \eqr{eq residual burgers}, is zero. 
\begin{equation}\label{eq residual burgers}
    r(u)=-u\frac{\partial u}{\partial x}
    +\nu\frac{\partial^2 u}{\partial x^2}-\frac{\partial u}{\partial t}.
\end{equation}

However, the same does not hold for $\bar{u}$. The ground truth for the closure term completely compensates the non-zero residual, i.e.  $\mathcal{C}^*=-r(\bar{u})$.

\subsubsection{Super-resolution} \label{appendix dataset super}
The snapshots for DG projection in this test are sliced from a public dataset~\citep{del2004scaling} for DNS solution for a channel flow at a friction Reynolds number $Re_{\tau}=\frac{u_\tau h}{2\nu}\approx {950}$, where $h$ is the channel height, $u_\tau = \sqrt{\tau/\rho}$ is the wall-friction velocity, defined on the averaged wall-friction $\tau$ and the density $\rho$. 

The slices are selected at different normalized wall-distances $z^+=z u_\tau/\nu$, where $z$ is the distance between the plane to the closer wall. 

\subsubsection{Rocket engine injector}\label{appendix dataset rocket}
The simulation for the public dataset~\citep{huang2019investigations} is performed using the finite-volume based General
Equation and Mesh Solver (GEMS)~\citep{harvazinski2015coupling}. 6 ms of flow is simulated in total at a time interval of $1 \times 10^{-4}$ ms. In our study the data is downsampled to an interval of $5 \times 10^{-4}$ ms.

\subsection{Network Training}\label{appendix network}
\subsubsection{Hyperparameters}
All models are trained with the Adam optimizer. Other training hyperparameters are summarized in Table.~\ref{table hyper}. For the closure models, the inputs $\mathbf{q}, \bar{u}$, and the output $\mathcal{C}$ are normalized by their respective maximum absolute values. No additional scaling used in the super-resolution task. For the reacting flow simulation task, the different variables in the input $\mathbf{q}$ are normalized to the same order of magnitude. The scaling coefficients are $C_{p}=5\times10^5, C_{u,v}=200, C_{T}=2500, C_Y=1$. For the output $\Delta \mathbf{q}$, the scaling coefficients are multiplied by an additional factor $C_\Delta=0.01$.

\begin{table}
\caption{Training hyperparameters.}
\label{table hyper}
\centering
\begin{tabular}{cc|ccc} 
\hline
\multicolumn{2}{c|}{Test case}                    & Batch size & Learning rate & Number of epochs  \\ 
\hline
\multicolumn{2}{c|}{Closure modeling}             & 1          & 0.001         & 300               \\
\multicolumn{2}{c|}{Super-resolution}             & 128        & 0.001         & 100               \\
\multirow{2}{*}{Reacting flow} & GNet-S/CP-GNet-S & 1          & 0.002         & 500               \\
                               & GNet-L/CP-GNet-L & 1          & 0.002         & 100               \\
\hline
\end{tabular}
\end{table}

\subsubsection{Training noise}
We follow the training noise injection strategy as in Refs.~\cite{pfaff2020learning, sanchez2020learning} to improve the robustness of prediction in the reacting flow simulation task. At the beginning of each training epoch, normally distributed noise $\epsilon\sim\mathcal{N}(0,0.0013^2)$ is added to the normalized inputs. The variance is selected based on the level of error in the prediction for one step at a time instance away from the training period. The source of this noise is assumed to be from the previous prediction step. The error is supposed to be compensated in the current prediction step; therefore, the noise is subtracted from the target output $\Delta\mathbf{q}$ after being scaled by $C_\Delta$.

\subsubsection{Adjustments for vertex-based graph}\label{appendix network vertex}

\subsection{Additional Analysis}\label{appendix additional}
\subsubsection{Closure modeling} \label{appendix additional closure}
\begin{figure}[!ht]
	\centering
	\includegraphics[width=0.45\columnwidth]{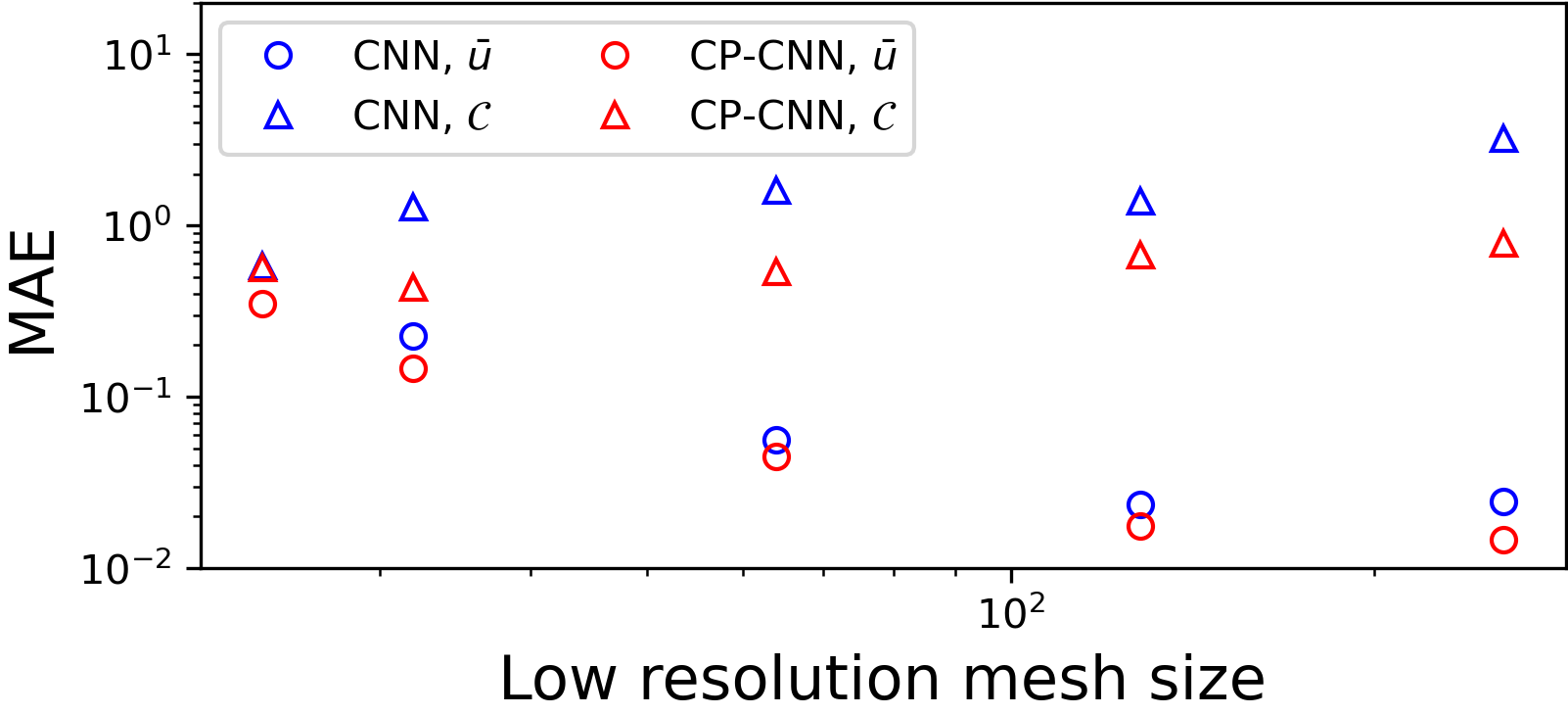}
	\caption{Average MAE for $\bar{u}$ (online) $\mathcal{C}$ (offline) under different low resolution mesh sizes. \textbf{The CNN model blows up at $\mathbf{n_x=24}$. The CP-CNN outperforms the CNN on all meshes.}}\label{fig closure widths}
\end{figure}

The comparison between CP-CNN and CNN is repeated on 4 other low resolution meshes of different sizes $n_x=\{24, 64, 128, 256\}$. The average MAE for the online computation for $\bar{u}$, and the offline single-step computation for $\mathcal{C}$ from the training IC is plotted in Fig.~\ref{fig closure widths}, along with the results for $n_x=32$ from Sec.~\ref{sec closure}. The CP-CNN outperforms the CNN on all meshes. Moreover, the CNN closure is unstable at the most coarse mesh, $n_x=24$, whereas the CP-CNN is stable.

\subsubsection{Super-resolution} \label{appendix additional super}
The definition for the energy spectra used in Sec.~\ref{sec superres} is given by:
\begin{align}
    e_x(k_x) = {\frac{1}{\pi}}\int_{-\infty}^{\infty}\left<u(x_0,y_0)u(x_0+x,y_0)\right>e^{-ik_xx}dx, \\
    e_y(k_y) = {\frac{1}{\pi}}\int_{-\infty}^{\infty}\left<u(x_0,y_0)u(x_0,y_0+y)\right>e^{-ik_yy}dy,
\end{align}
where $\left<\cdot\right>$ denotes the average over homogeneous directions, which is the entire plane in this case. Similar to the power spectral density for a time series that describes the energy distribution over different frequencies, the energy spectra describes the energy distribution of a spatial field over different wave-numbers $k={2\pi/\lambda}$, $\lambda$ being the wavelength. 

\subsubsection{Rocket engine injector} \label{appendix additional rocket}
\begin{figure}[!ht]
	\centering
	\includegraphics[width=1\columnwidth]{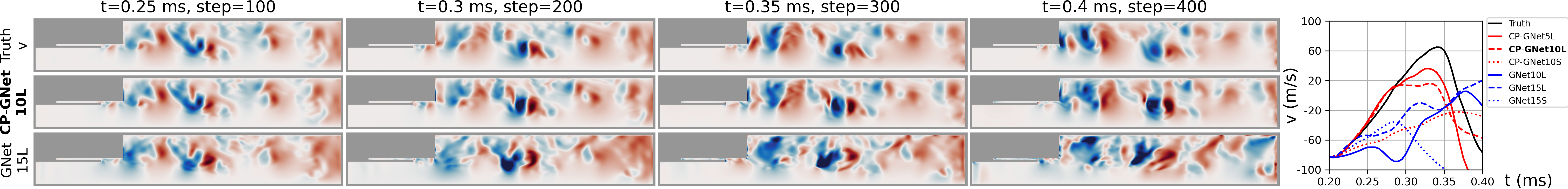}
	\includegraphics[width=1\columnwidth]{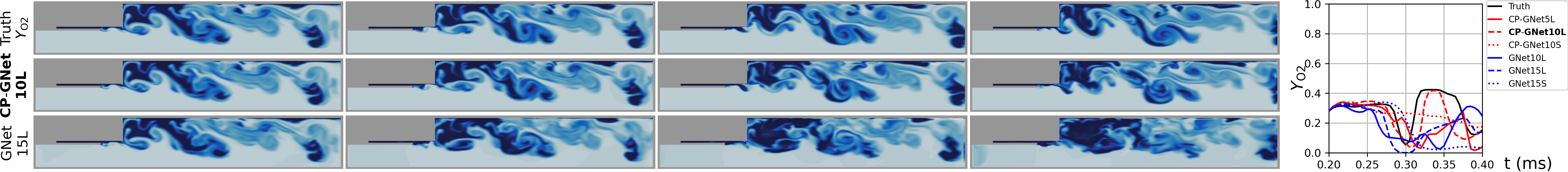}
	\includegraphics[width=1\columnwidth]{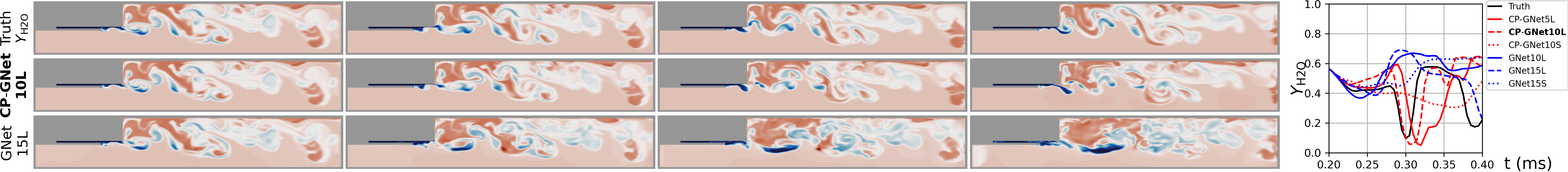}
	\includegraphics[width=1\columnwidth]{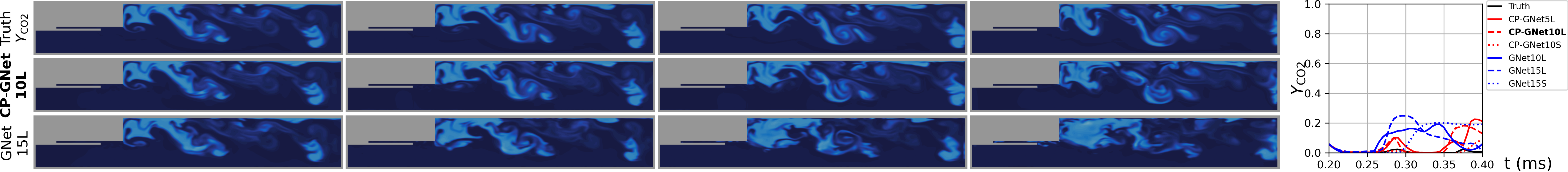}
	\caption{Predictions for the rest variables $v$, $Y_{\mathrm{O2}}$, $Y_{\mathrm{H2O}}$, $Y_{\mathrm{CO2}}$.}\label{fig deepblue rest}
\end{figure}

\paragraph{Additional variables.} As a supplement for Fig.~\ref{fig deepblue}, predictions for the rest variables, $v$, $Y_{\mathrm{O2}}, Y_{\mathrm{H2O}}, Y_{\mathrm{CO2}}$, are provided in Fig.~\ref{fig deepblue rest}. They further validate the conclusions in Sec.~\ref{sec deepblue}. 

\begin{figure}[!ht]
	\centering
	\includegraphics[width=1\columnwidth]{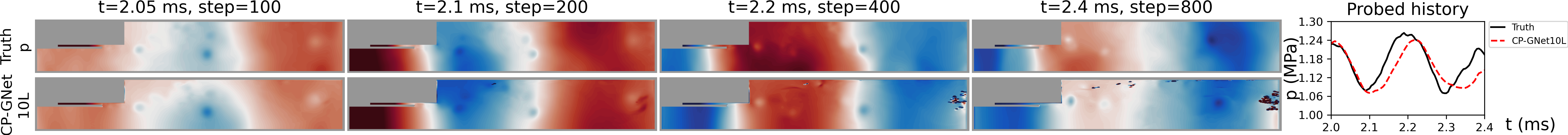}
	\includegraphics[width=1\columnwidth]{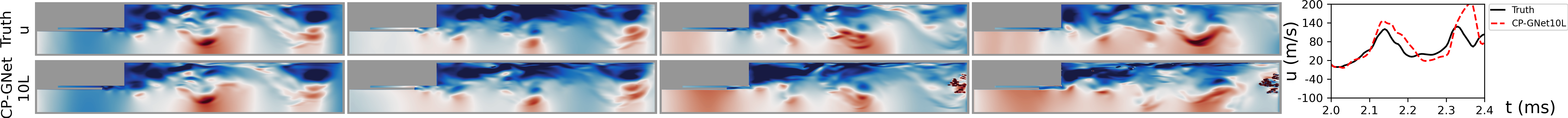}
	\includegraphics[width=1\columnwidth]{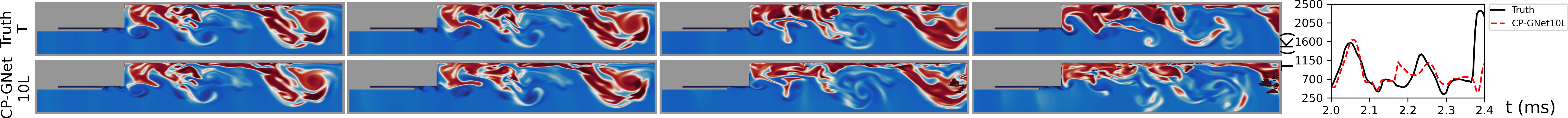}
	\includegraphics[width=1\columnwidth]{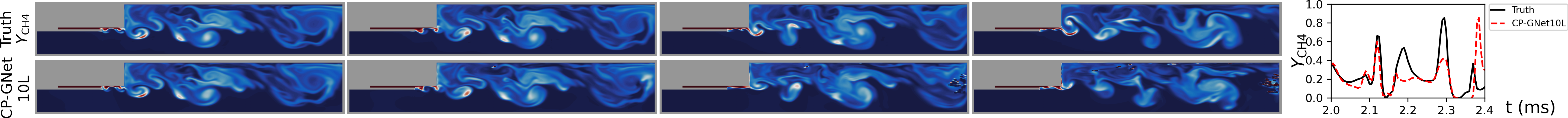}
	\caption{Predictions from a new time instance $t=2$ ms for a longer period of 0.4 ms. The initial 0.15 ms shows a similar level of accuracy as the appended case. The prediction becomes unstable in the long term, illustrated by scattered extreme values in the contours.}\label{fig deepblue long}
\end{figure}

\paragraph{Prediction for a distanced and longer period.} To test generalization performance, the CP-GNet10L model is used to perform prediction at a new time instance $t=2$ ms, which is away from the training period. The predicted period is also doubled to 0.4 ms. The results are provided in Fig.~\ref{fig deepblue long}. For the first 0.15 ms, a similar level of accuracy is obtained compared with the previous run that is appended to the end of the training period. However, the prediction is unstable in the long-term, illustrated by scattered extreme values in the contours. The authors acknowledge that long-time stability guarantees is a major limitation of the current - and existing - work in the domain of data-driven flow predictions.

\begin{figure}[!ht]
	\centering
	\includegraphics[width=1\columnwidth]{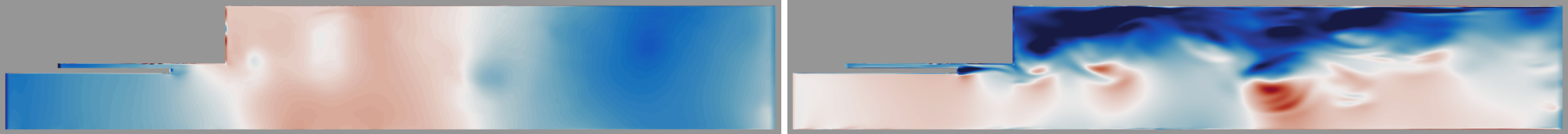}
	\caption{Results on a graph without ghost edges for $p$ (left) and $u$ (right). The accumulation of error in the near-wall regions is clearly visible.}\label{fig deepblue ghost}
\end{figure}

\paragraph{Results without the boundary treatment.} The CP-GNet10L model is re-trained on a graph without ghost edges for the wall boundaries. The prediction is again started at the end of the training ($t=0.2$ ms). The results for $p$ and $u$ at the last prediction step $t=0.4$ ms are shown in Fig.~\ref{fig deepblue ghost}. The accumulation of error is clearly visible in several near-wall regions, which validates our proposed boundary treatment. 

\paragraph{Timing.} The prediction for 0.1 ms of flow with the CP-GNet10L model takes 53 seconds on one Nvidia RTX A6000 GPU, or 599 seconds on 40 CPU cores. In comparison, the original simulation for 6 ms of flow takes approximately 1200 CPU hours~\citep{mcquarrie2020data}. Due to different hardware configurations, no direct comparison can be made. However, we can safely estimate a 2.5x$\sim$3x speedup on CPUs, and a 25x$\sim$30x speedup when a GPU is utilized. 
\section{Generalizable \& Exact Fitting with CP}\label{sec appendix b}
\begin{figure}[!ht]
\captionsetup[subfigure]{justification=centering} 
	\centering
    \hspace*{\fill}
	\subfloat[{$\Delta x=0.03, \Delta y=0.02,$ \protect\\ $\mathbf{a}=[1, -0.8]^T, \nu=0.035$}\label{fig adv grid}]{
	\begin{minipage}{0.31\linewidth}
	\centering
	\includegraphics[width=1\columnwidth]{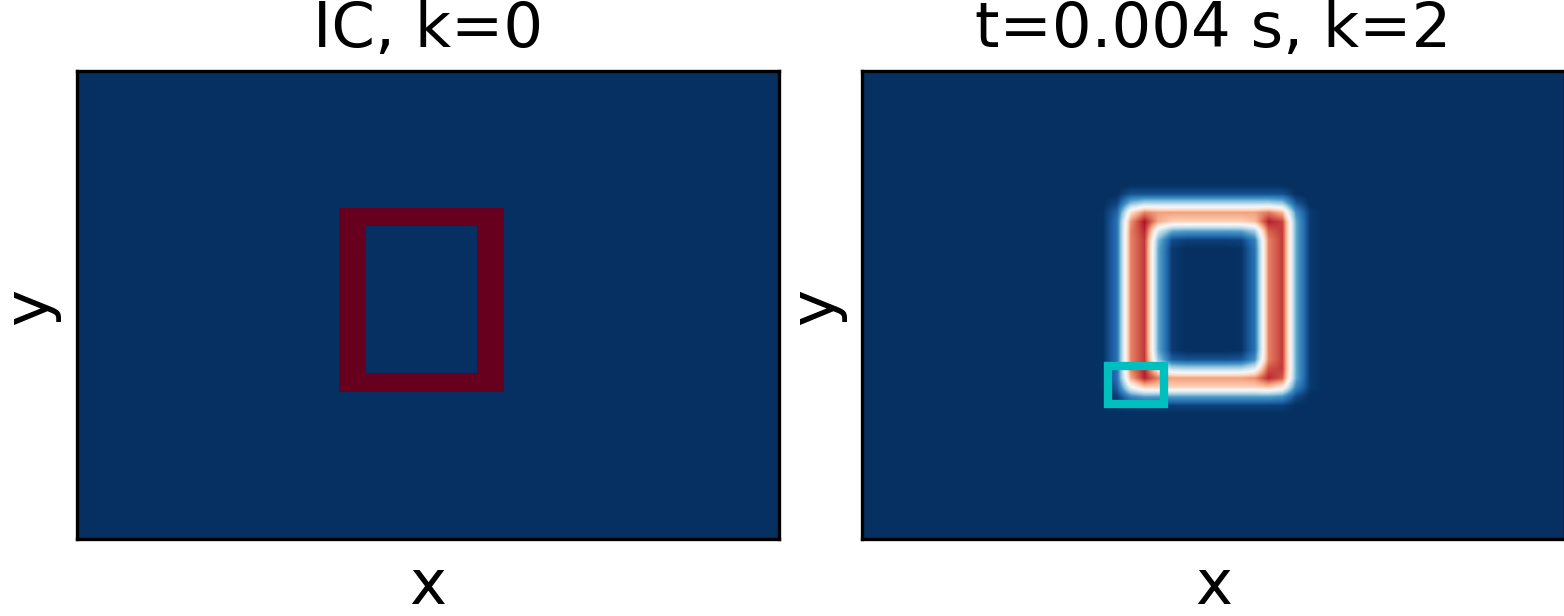}\\
	\hfill\includegraphics[width=0.8\columnwidth]{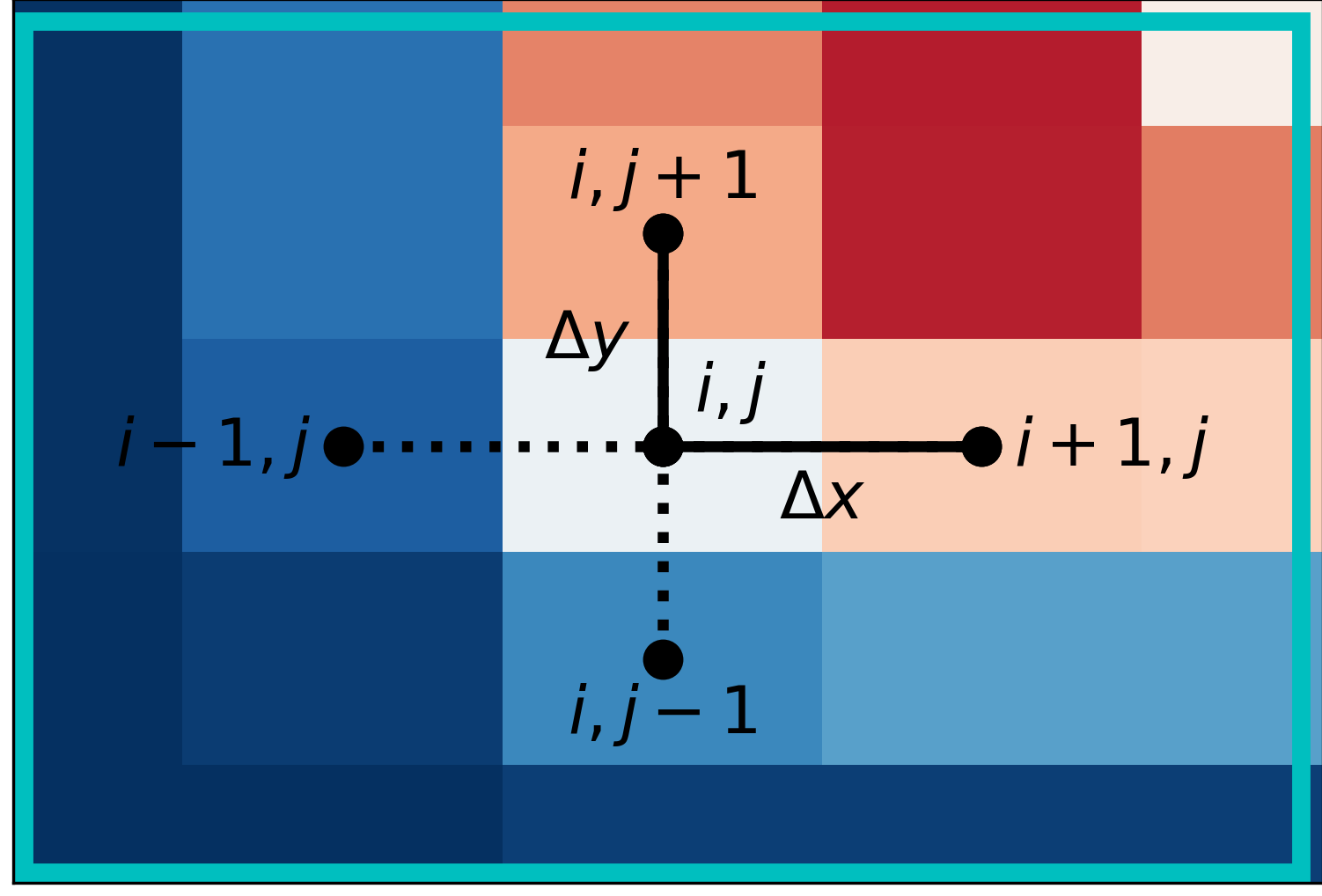}
	\end{minipage}
	}
    \hspace*{\fill}
	\subfloat[{$\Delta x=0.02, \Delta y=0.02,$ \protect\\ $\mathbf{a}=\left[1.2, 1.2\right]^T, \nu=0.035$}]{
	\begin{minipage}{0.31\linewidth}
	\centering
	\includegraphics[width=1\columnwidth]{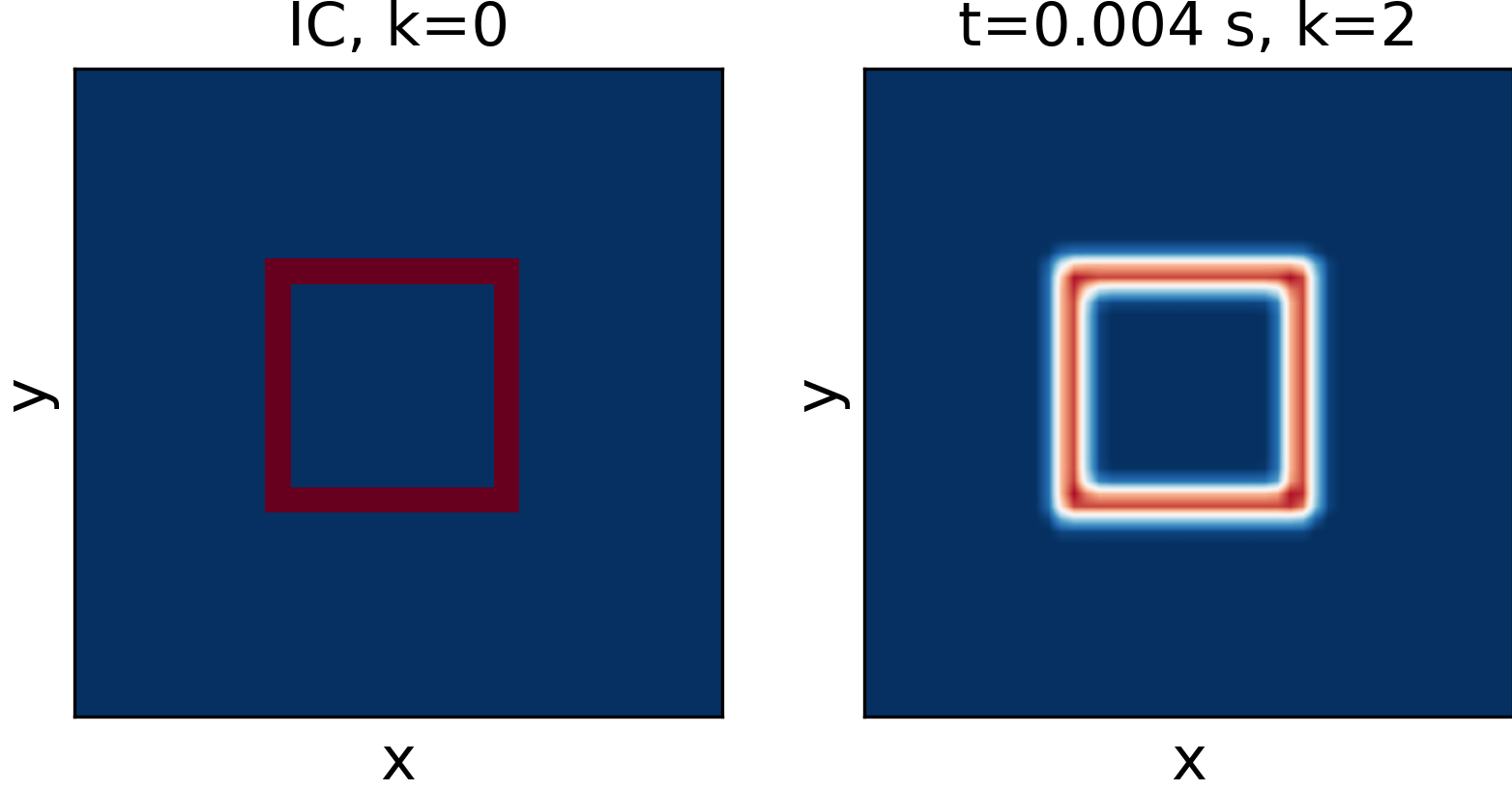}
	\end{minipage}
    }
    \hspace*{\fill}
	\subfloat[{$\Delta x=0.02, \Delta y=0.024,$ \protect\\ $\mathbf{a}=[-1, 1]^T, \nu=0.04$}]{
	\begin{minipage}{0.31\linewidth}
	\centering
	\includegraphics[width=1\columnwidth]{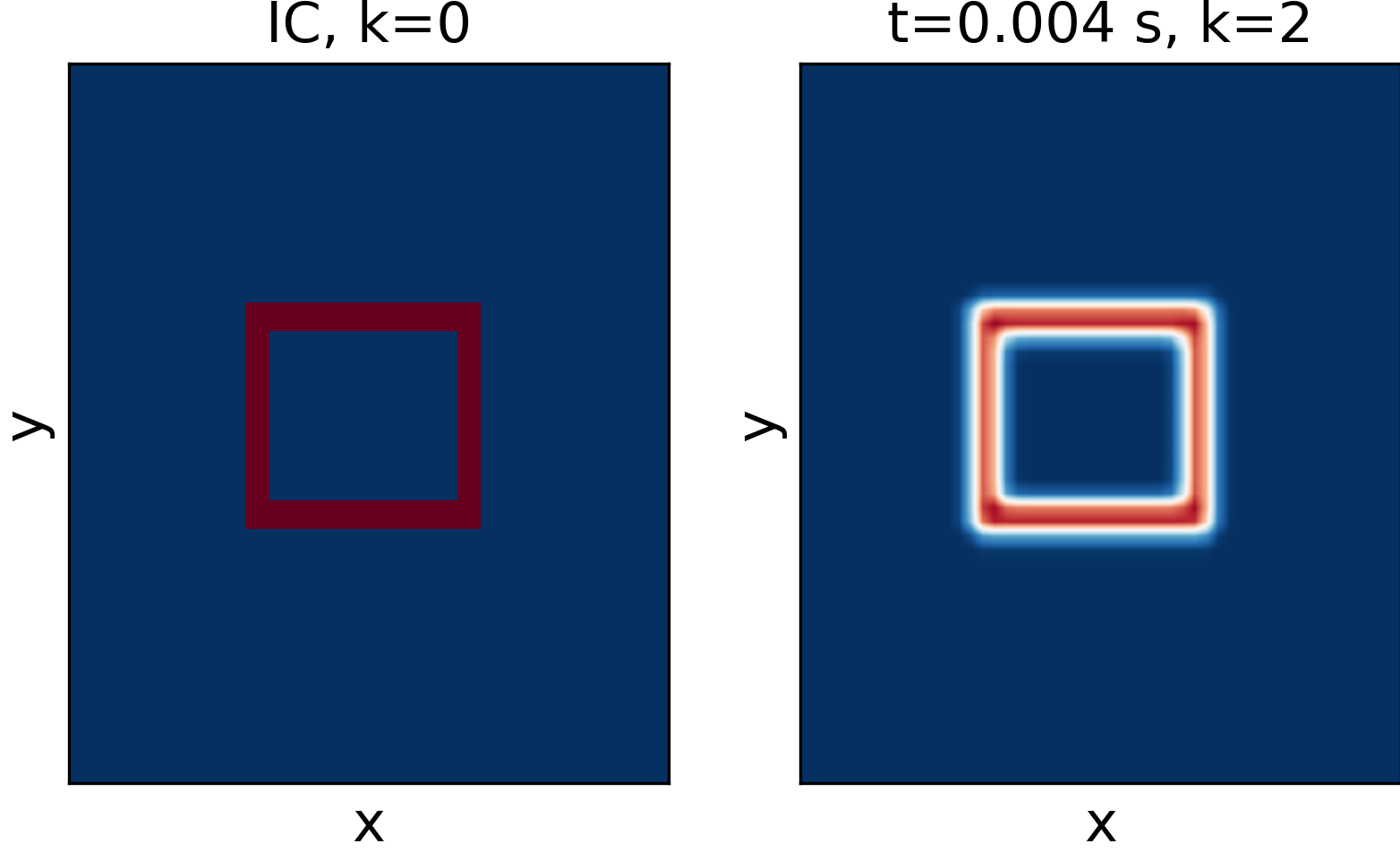}
	\end{minipage}
    }
    \hspace*{\fill}
	\caption{Training data. \textbf{Each case consists of only the IC and two solution steps.}}\label{fig adv train}
\end{figure}

In this appendix we demonstrate - as a proof-of-concept problem - how  discretized PDE terms can be fitted exactly with simple CP layers in the solution of the 2D advection-diffusion equation. This experiment is performed in the limit of extremely sparse data snapshots. With periodic boundary conditions, the governing equations are:

\begin{equation}\label{eq advdiff}
\begin{split}
    &\frac{\partial u(x,y,t)}{\partial t} + \mathbf{a}\nabla u(x,y,t) - \nu\nabla^2u(x,y,t)=0,\\
    &x\in[0,W], y\in[0,H], t\in[0,T],\\
    &u(0,y,t)=u(W,y,t), u(x,0,t)=u(x,H,t),
\end{split}
\end{equation}
where $\mathbf{a}=[a_x, a_y]^T$ is the advection velocity vector. The initial condition (IC) has a rectangular frame with $u=1$ in the center, and $u=0$ elsewhere, as shown in Fig.~\ref{fig adv train}.

The first-order upwind discretization of \eqr{eq advdiff} is given by:
\begin{equation}\label{eq advdiff discrete}
\begin{split}
    &\frac{\Delta u^k_{i,j}}{\Delta t}+
    \frac{a_x+|a_x|}{2\Delta x}(u^k_{i,j}-u^k_{i-1,j})+\frac{a_x-|a_x|}{2\Delta x}(u^k_{i+1,j}-u^k_{i1,j})+
    \frac{a_y+|a_y|}{2\Delta y}(u^k_{i,j}-u^k_{i,j-1})\\
    &+\frac{a_y-|a_y|}{2\Delta y}(u^k_{i,j+1}-u^k_{i,j})
    -\nu\frac{u^k_{i-1,j}-2u^k_{i,j}+u^k_{i+1,j}}{\Delta x^2}
    -\nu\frac{u^k_{i,j-1}-2u^k_{i,j}+u^k_{i,j+1}}{\Delta y^2}=0,
\end{split}
\end{equation}
where $i$ and $j$ are the grid indices in the $x$ and $y$ directions, respectively, and $\Delta x$ and $\Delta y$ are the distances between grid points, as illustrated in Fig.~\ref{fig adv grid}.

Inspired by \eqr{eq advdiff discrete}, a neural network model is constructed  using a dense layer and two 2D CP-Convolution (CP-Conv) layers in the form:
\begin{equation}\label{eq advdiff nn}
\begin{split}
    \mathbf{h} = \mathrm{ReLU}(\mathbf{W}_1\mathbf{a}),\\
    \Delta \mathbf{u}^k=
    \left<\mathbf{W}_2,\frac{\Delta t}{\Delta x}\mathbf{h}\right>*\mathbf{u}^k+\left(\left<\mathbf{W}_2,\frac{\Delta t}{\Delta y}\mathbf{h}\right>*(\mathbf{u}^k)^T\right)^T\\
    +(\mathbf{W}_3\frac{\nu\Delta t}{\Delta x^2})*\mathbf{u}^k+((\mathbf{W}_3\frac{\nu\Delta t}{\Delta y^2})*(\mathbf{u}^k)^T)^T,
\end{split}
\end{equation}
where $\mathbf{W}_1\inr{2\times2}$ is the weight for the dense layer, and $\mathbf{h}$ is the hidden output. $\mathbf{W}_2\inr{3\times1\times1\times2}$ is weight for the first CP-Conv kernel, taking $\frac{\Delta t}{\Delta x}\mathbf{h}$ or $\frac{\Delta t}{\Delta y}\mathbf{h}$ as the condition parameter. $\mathbf{W}_3\inr{3\times1\times1\times1}$ is weight for the second CP-Conv kernel, taking $\frac{\nu\Delta t}{\Delta x^2}$ or $\frac{\nu\Delta t}{\Delta y^2}$ as the condition parameter. $(\cdot*\cdot)$ denotes the convolution operation.

In common deep learning applications,  a large amount of training data is required to train the model sufficiently, and to avoid overfitting. In this test case, however, we assess the ability of the model to approximate the truth at a machine precision level using limited data. As a demonstration, we use only 3 sets of 2-step training data snapshots, each for a different set of parameters $\{\Delta x, \Delta y, \mathbf{a}, \nu\}$ to train a model represented by \eqr{eq advdiff nn}. Each set only has solutions for two time steps beyond the initial condition. All of the training cases are present in Fig.~\ref{fig adv train}. The model is trained with Adam optimizer with the training hyperparameters listed in Table~\ref{table advdiff}.

The weights learnt~\footnote{$\mathbf{W}_2$ and $\mathbf{W}_3$ squeezed for simplicity} are:
{\scriptsize
$$\mathbf{W}_1=\begin{bmatrix} 0.33237486 &0 \\ 0 &-0.5253752 \end{bmatrix},
\mathbf{W}_2=\begin{bmatrix} 3.00869074 &-3.00892553 &-8.69012577\times10^{-5} \\ 2.38949641\times10^{-5} &-1.90361486 &1.90334544 \end{bmatrix}^T,$$
$$\mathbf{W}_3=\left[0.99999235, -1.99994342,  1.00001345\right]^T.$$
}

Indeed when the weights are substituted into \eqr{eq advdiff nn}, we recover \eqr{eq advdiff discrete} to 4-decimal-point precision, as the ideal weight combination is 
{\scriptsize
$$\mathbf{W}_1=\begin{bmatrix} 0.5c_1 &0 \\ 0 &-0.5c_2 \end{bmatrix}, \mathbf{W}_2=\begin{bmatrix} 1/c_1 &-1/c_1 &0 \\ 0 &-1/c_2 &1/c_2 \end{bmatrix}^T, \mathbf{W}_3=\left[1,-2,1\right]^T,$$}
where $c_1, c_2 \in \mathds{R}_{\ne 0}$.


\begin{table}
\caption{Training hyperparameters.}
\label{table advdiff}
\small
\centering
\begin{tabular}{c|c} 
\hline
Number of snapshots & 6\\
Grid points per snapshot & $51\times51$\\
Batch size & 1\\
Initial learning rate & 0.1\\
Final learning rate & 0.0003\\
Number of epochs  & 10000\\ 
\hline
\end{tabular}
\end{table}

\begin{figure}[!ht]
	\centering
	\includegraphics[width=1\columnwidth]{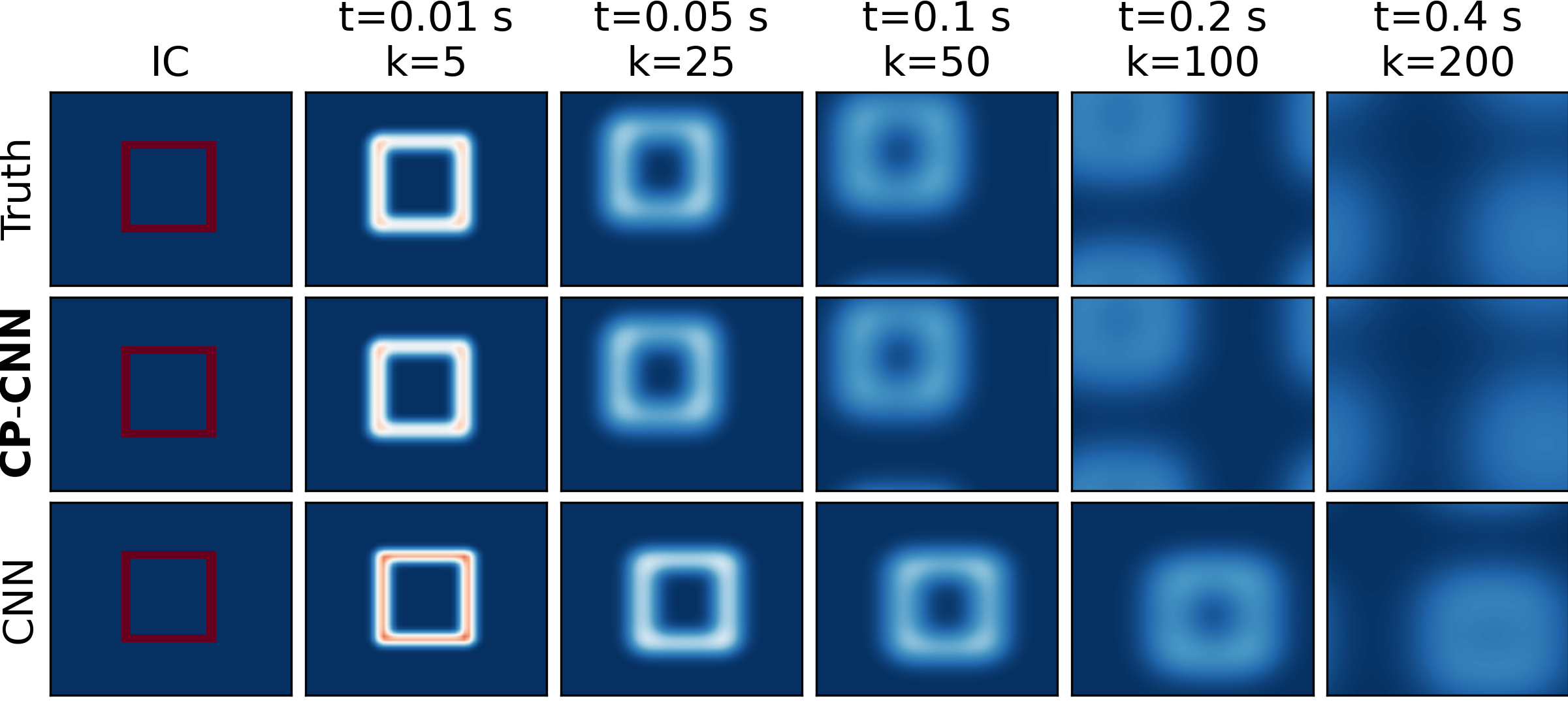}
	\caption{Prediction results. CP-CNN fits the discretized model exactly. The CNN is highly inaccurate.}\label{fig adv test}
\end{figure}

The model is used to perform rollout prediction for 200 steps at a new initial condition and set of parameters outside the training range, $\Delta x=0.02, \Delta y=0.016, \mathbf{a}=[-1.5, 1.5]^T, \nu=0.02$. The result is present in Fig.~\ref{fig adv test}. The L1 error is less than \num{1e-4}. 

In comparison, a CNN with the CP-Conv layers in \eqr{eq advdiff nn} replaced by standard convolution layers are applied to the same task. It can be seen from Fig.~\ref{fig adv test} that the CNN can represent neither the advection nor the diffusion correctly.

It is thus clear that for this class of problems, the utility of conventional deep learning is questionable. The authors note that, in practical and more complex applications, exact fitting will not be achievable even with CP. However, as an idealized demonstration, this test serves the purpose to show how CP networks can represent functional relations between parameters and variables to reduce the training effort for certain terms, to improve accuracy, and offer generalizable predictions.
\section{Comparison against MeshGraphNets}\label{appendix comparison}
Although both methods are designed for mesh-based simulations using graphs, the CP-GNet is inspired by the finite volume method, leading to different graph representations. In addition to the conditional parametrization, the three main differences between the two approaches  are: 1) The edges are unidirectional between nodes for the CP-GNet v.s. bidirectional for the MeshGraphNets; 2) Graph nodes are located at mesh cell centers for the CP-GNet v.s. mesh vertices for the MeshGraphNets;  3) The boundary conditions are implemented by adding ghost edges for the CP-GNet v.s. distinguishing node labels for the MeshGraphNets. Due to these differences, a strict direct comparison between the two methods becomes infeasible. To provide a meaningful comparison, both models are tested on a FVM dataset (reacting flow) and a FEM dataset ( incompressible flow over a cylinder). Minimal adjustments to the models/input features are made accordingly for migration purposes.

\subsection{Reacting flow}
The long-training-period experiment setting from Sec.~\ref{sec deepblue} is used. To apply the MeshGraphNets, one-hot labels distinguishing fluid cells and different types (symmetric wall/no-slip wall/inlet/outlet) of boundary cells, as well as the cell volumes are added to the node features. Face areas between cells are added to the edge features. Moreover, we also tested a wider version of the MeshGraphnets, with the default 128-unit MLPs replaced by 256-unit ones, due to the complex reaction physics in this task. Both MeshGraphNets are trained using the same training hyper-parameters as the CP-GNet10L model from Sec.~\ref{sec deepblue}, and compared with the latter.

\begin{figure}[!ht]
	\centering
	\includegraphics[width=1\columnwidth]{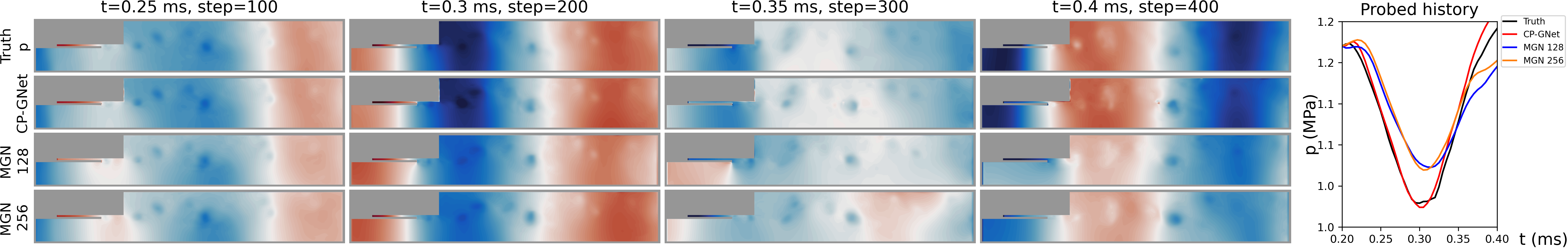}
	\includegraphics[width=1\columnwidth]{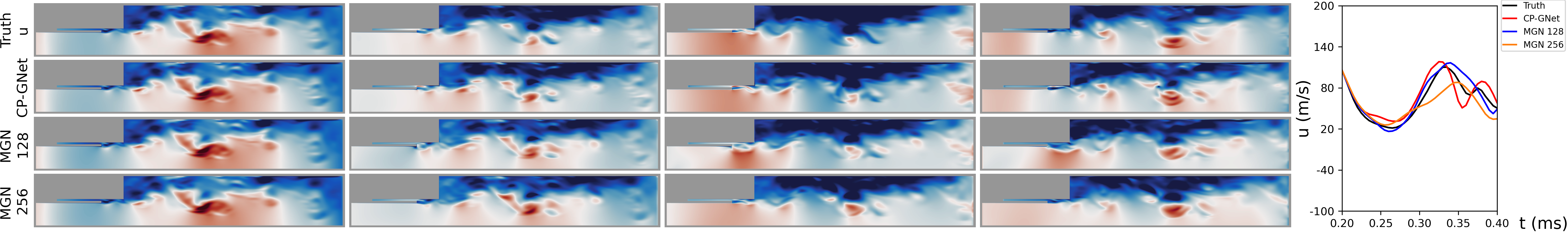}
	\includegraphics[width=1\columnwidth]{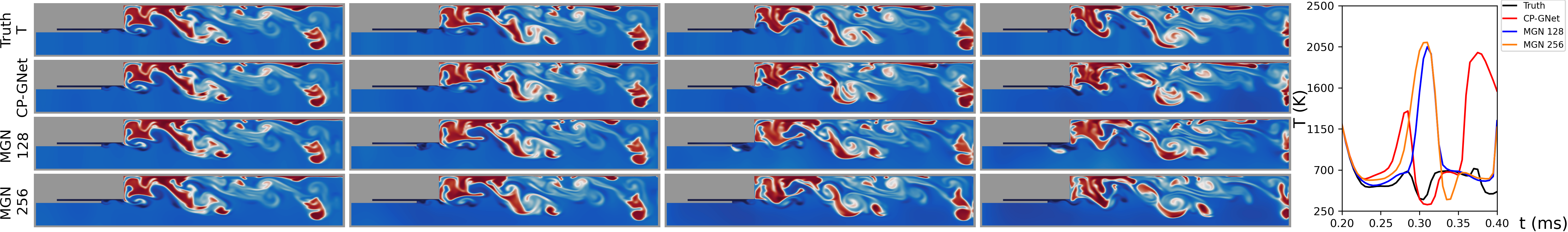}
	\includegraphics[width=1\columnwidth]{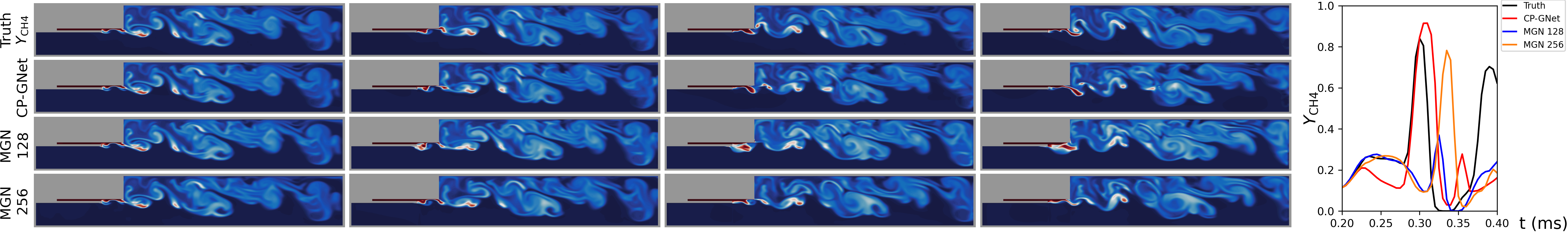}
	\caption{Predicted reacting flow. For each variable from top to bottom: ground truth, CP-GNet, 128-unit MeshGraphNets (MGN 128), 256-unit MeshGraphNets (MGN 256).}\label{fig comparison deepblue}
\end{figure}

Evaluations are again performed on the representative variables $p,u,T, Y_{\mathrm{CH4}}$. The predicted flow fields are visualized in Fig.~\ref{fig comparison deepblue}. It can be seen that the CP-GNet is visually closer to the truth, especially in the phases of the probed peaks. The averaged inference time and RMSE for the normalized variables (with mean subtracted, divided by standard deviation) at different rollout steps is reported in Table~\ref{table comparison deepblue}. The present model provides a lower RMSE throughout the prediction. 

\begin{table}
\caption{Averaged inference time and RMSE for reacting flow.}\label{table comparison deepblue}
\small
\begin{tabular}{l|cccc}
\hline
Model                    & \begin{tabular}[c]{@{}c@{}}Time/step\\ ms\end{tabular} & \begin{tabular}[c]{@{}c@{}}RMSE 1-step\\ $\times 10^{-3}$\end{tabular} & \begin{tabular}[c]{@{}c@{}}RMSE rollout-50\\ $\times 10^{-3}$\end{tabular} & \begin{tabular}[c]{@{}c@{}}RMSE rollout-all\\ $\times 10^{-3}$\end{tabular} \\ \hline
CP-GNet & 261 & 0.29 & 6.8 & 46.1 \\
128-unit MeshGraphNets & 203 & 0.42 & 10.4 & 62.8 \\
256-unit MeshGraphNets & 296 & 0.41 & 10.8 & 58.4 \\ \hline
\end{tabular}
\end{table}

\subsection{Incompressible flow over a cylinder}
For this experiment, the setting for  incompressible flow over a cylinder from the MeshGraphNets~\citep{pfaff2020learning} is adopted. The task is to predict the 2D velocity components stored on the vertices on irregular triangular meshes. The data includes 1000 training trajectories and 100 testing ones, each with 600 steps. In the original setting, 10 million training steps are used, which takes several days on a single GPU. In this work, an additional comparison is performed after 2.5 million training steps to evaluate the training efficiency of the models.

The MeshGraphNets results are generated using the official code. The CP-GNet model is migrated to the same pipeline, with two minor adjustments made: 1) A two-layer MLP is added to the node encoder to process the node label; The network width is increased from 36 to 64 to process the additional features. 2) The CP-Dense layer in the ``source term'' section in the processor is removed as there is no chemical reaction taking place in this test.

\begin{figure}[!ht]
	\centering
	\includegraphics[width=1\columnwidth]{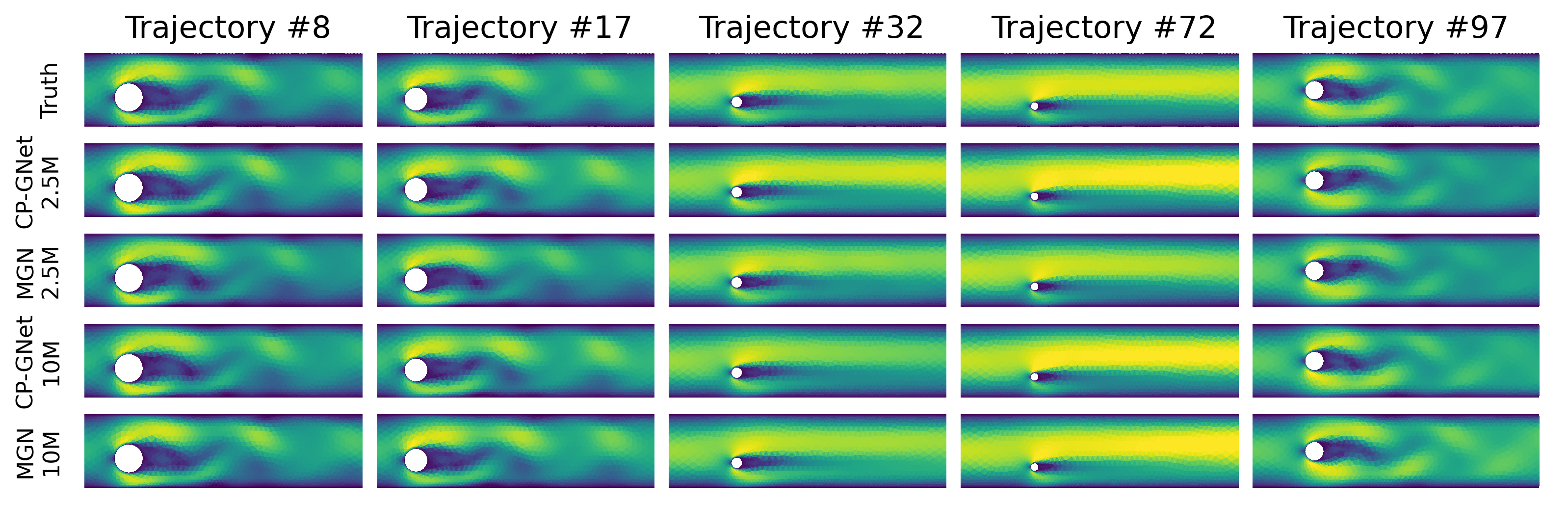}
	\caption{Velocity magnitude for the last step in the rollout prediction for random testing trajectories. From top to bottom: ground truth, CP-GNet, MeshGraphNets (MGN).}\label{fig comparison cylinder}
\end{figure}

The averaged inference time and RMSE for the testing trajectories are summarized in Table~\ref{table comparison cylinder}. It should be noted that a single A6000 GPU is used in our tests, and a V100 is used in \citep{pfaff2020learning}, and noticeably different numbers are reported. At a smaller number of training steps (2.5 M), our model provides a higher single-step RMSE, yet a lower long-rollout RMSE. Compared with the previous test, the efficiency of our model is largely affected by the additional processing of node labels and a larger network width, running significantly slower than the MeshGraphNets. The predicted final steps for 5 randomly selected testing trajectories are visualized in Fig.~\ref{fig comparison cylinder}. At 2.5 M training steps,  the CP-GNet performs  better in two unsteady cases (trajectories \#8 and \#17). However, it over-predicts the velocity magnitude in the two steady cases (trajectories \#32 and \#72). Meanwhile, the MeshGraphNet under-predicts in one of them (trajectory \#72). At the larger number of training steps (10 M), the MeshGraphNets shows a lower error across the prediction period, with the gap between the two models decreasing with the number of rollout steps. The differences in the final step predictions become less significant. 

It should be pointed out that between this and the previous test problems, there are many factors that can lead to changes in model performances, including the type of the ground truth solver and data (cell-centered FVM vs. vertex-centered FEM), the number of variables (8 vs. 2), the implement of boundary conditions (ghost edges vs. node labels). Based on these specific results, one cannot make a definitive statement on the relative merits of each of the two approaches.

\begin{table}
\caption{Averaged inference time and RMSE for flow over cylinder.}\label{table comparison cylinder}
\small
\begin{tabular}{l|cccc}
\hline
Model (training steps) & \begin{tabular}[c]{@{}c@{}}Time/step\\ ms\end{tabular} & \begin{tabular}[c]{@{}c@{}}RMSE 1-step\\ $\times 10^{-3}$\end{tabular} & \begin{tabular}[c]{@{}c@{}}RMSE rollout-50\\ $\times 10^{-3}$\end{tabular} & \begin{tabular}[c]{@{}c@{}}RMSE rollout-all\\ $\times 10^{-3}$\end{tabular} \\ \hline
CP-GNet (2.5 M) & 16 & 3.3 & 12.4 & 62.5 \\
CP-GNet (10 M) & 16 & 2.8 & 9.9 & 54.0 \\
MeshGraphNets (2.5 M, tested)   & 9 & 2.1 & 8.7 & 68.5 \\
MeshGraphNets (10 M, tested)   & 9 & 1.9 & 6.9 & 50.1 \\
MeshGraphNets (10 M, reported~\citep{pfaff2020learning}) & 21 & $2.34 \pm 0.12$ & $6.3 \pm 0.7$ & $40.88 \pm 7.2$ \\ \hline
\end{tabular}
\end{table}

\begin{figure}[!ht]
	\centering
	\includegraphics[width=0.9\columnwidth]{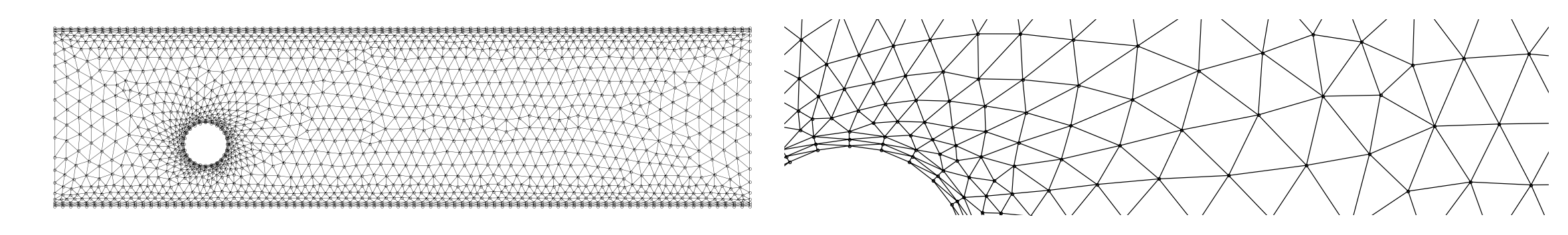}
	\caption{Example irregular mesh for the flow over cylinder with a zoomed-in view on the right}\label{fig grid cylinder}
\end{figure}

\end{document}